\documentclass{article}

\usepackage[preprint]{neurips_2020}

\usepackage[utf8]{inputenc} % allow utf-8 input
\usepackage[T1]{fontenc}    % use 8-bit T1 fonts
\usepackage{hyperref}       % hyperlinks
\usepackage{url}            % simple URL typesetting
\usepackage{booktabs}       % professional-quality tables
\usepackage{amsfonts}       % blackboard math symbols
\usepackage{nicefrac}       % compact symbols for 1/2, etc.
\usepackage{microtype}      % microtypography

\usepackage{amsmath}
\usepackage{mathtools}
\usepackage{bm}
\usepackage{bbm}
\usepackage{amssymb} 
\usepackage{graphicx} 
\usepackage[UKenglish]{babel}
\usepackage[export]{adjustbox}
\usepackage{multirow}
\usepackage{enumerate}
\usepackage{tabularx} 
\usepackage{fancyhdr} 
\usepackage{xcolor}

\setcitestyle{numbers,square,comma}

\newcommand{\IR}{\mathbb{R}}

\newcommand{\IN}{\mathbb{N}}

\DeclareMathOperator\BN{BN}
\DeclareMathOperator\MHA{MHA}
\DeclareMathOperator\MHAres{MHA^{res}}
\DeclareMathOperator\FF{FF}

\DeclareMathOperator\FFres{FF^{res}}

\DeclareMathOperator\SA{SA}
\title{Learning to Solve Vehicle Routing Problems with Time Windows through Joint Attention}

\author{
	Jonas K.~Falkner\\	
	Department of Computer Science\\
	University of Hildesheim\\
	31141 Hildesheim, Germany \\
	\texttt{falkner@ismll.uni-hildesheim.de} \\
	\And
	Lars Schmidt-Thieme \\
	Department of Computer Science\\
	University of Hildesheim\\
	31141 Hildesheim, Germany \\
	\texttt{schmidt-thieme@ismll.uni-hildesheim.de} 
}

\begin{document}

\maketitle

% abstract
%%%%%%%%%%%%%%%%%%%%
\begin{abstract}
Many real-world vehicle routing problems involve rich sets of constraints with
respect to the capacities of the vehicles, time windows for customers etc. While
in recent years first machine learning models have been developed to solve basic
vehicle routing problems faster than optimization heuristics, complex constraints
rarely are taken into consideration. Due to their general procedure to construct
solutions sequentially route by route, these methods generalize unfavorably to such
problems. In this paper, we develop a policy model that is able to start and 
extend multiple routes concurrently
by using attention on the joint action space of several tours. In that way the model 
is able to select routes and customers and thus learns to make difficult trade-offs between routes. 
In comprehensive experiments on three variants of the vehicle routing problem with time windows 
we show that our model called JAMPR works well for
different problem sizes and outperforms the existing state-of-the-art constructive model.
For two of the three variants it also creates significantly better solutions than a comparable meta-heuristic solver. 
\end{abstract}

% content
%%%%%%%%%%%%%%%%%%%%
%%%%%%%%%%%%%%%%%%%%%%%%%%%%%%%%%%%%%%%%%

\section{Introduction} \label{sec:introduction}
% several types of CVRP-TW (wait, soft, hard)...
%This paper is concerned with the well known extension of the Capacitated Vehicle Routing Problem (CVRP) which includes additional time windows (TW). 
The standard CVRP is a NP-hard combinatorial optimization problem which consists of finding the optimal set of tours for a fleet of capacitated vehicles which have to serve the varying demand of multiple customers. %All vehicles start at a specific depot node to which they have to return at the end of their tour. 
However, for many practical scenarios the problem model of the CVRP is oversimplified and leads to highly sub-optimal solutions since it only takes distances into account. Therefore an important extension introduces an additional time dimension with respective time windows. 
The CVRP-TW consequently has a high relevance for many practical applications and has been a subject of study for several decades in the Operations Research community \citep{solomon1987algorithms, braysy2005vehicle, toth2014vehicle}. 
%While the solution space of the CVRP only has restrictions regarding the vehicle capacity and the customer demands, solutions to the CVRP-TW need to respect additional restrictions regarding the time a vehicle arrives at a customer node, the duration of the service task fulfilling the customer demand and the respective travel times. This complicates the construction of a solution in a sequential fashion.

Since the seminal work of \citet{vinyals_pointer_2015} showing that deep learning methods are a valid approach to solve combinatorial optimization problems, the field has evolved %seen several innovations 
in recent years. 
Apart from general graph related problems like maximum cut \citep{khalil_learning_2017} or link prediction and graph partitioning \citep{wilder2019end}, the Traveling Salesman Problem (TSP) %as the simplest routing problem 
was most extensively covered \citep{bello2016neural, khalil_learning_2017, deudon_learning_2018, fu_targeted_2019, xing_graph_2019}. 
Meanwhile however, there exist also several approaches to solve VRPs \citep{nazari_reinforcement_2018, kool2018attention, chen2019learning, lu_learning_based_2019}.
%In contrast, %to the best of our knowledge, the CVRP only has been tackled by four works so far 
%the CVRP has received much less attention \citep{nazari_reinforcement_2018, kool2018attention, chen2019learning, lu_learning_based_2019}.
%Furthermore up to date there does not exist any published attempt to directly solve the much more complex CVRP-TW albeit its extensive coverage in the OR literature. 

In the same way as traditional heuristics for routing problems can be categorized as \textit{construction} or \textit{improvement} heuristics \citep{toth2014vehicle}, respective learned methods exhibit some similar characteristics. While the first approaches of \citet{nazari_reinforcement_2018} and \citet{kool2018attention} \textit{construct} a solution to the CVRP sequentially one node at a time, other work \citep{chen2019learning, lu_learning_based_2019} focuses on iteratively \textit{improving} an existing solution. 
%Both approaches have their advantages and disadvantages. 
In general improvement approaches assume an existing start solution which they can iteratively improve. While such an initial solution might be easily found for simple problems like the TSP or CVRP, this is increasingly difficult for more constrained problems. In fact just finding a first feasible solution for the CVRP-TW given a fixed number of vehicles is a NP-hard problem all by itself \citep{savelsbergh1985local}. 
Another shortcoming of improvement approaches lies in their expensive iterative nature. Compared to construction methods which can produce a useful %(if not necessarily optimal) 
solution in one forward pass within seconds, iterative improvement requires a not negligible number of iterations to achieve a suitable performance. 
This number grows relative to the problem complexity, number of customers $N$ and the quality of the initial solution while constructive approaches always require only $N$ steps for greedy decoding. Finally an improvement method can always be used on top of the solution produced by constructive approaches.
%In contrast, one criticism regarding construction methods is that they are not able to revisit earlier decisions in the construction procedure possibly leading to local optima and consequently sub-optimal solutions. This problem however can mostly be resolved by employing different beam search or sampling methods. Another
One drawback of existing sequential construction approaches in case of the CVRP-TW is the very limited information on which the next decision of the agent is based. 
%For example in case of the Attention Model (AM) of \citet{kool2018attention} the context for each decoding step only includes the graph embedding, the remaining capacity and the last node visited. 
While our results show that this is of minor concern for the CVRP, the highly restricted solution space of the CVRP-TW leads to large inefficiencies when tours are created sequentially without any information about the %position and load of the other vehicles 
other tours at a given time. 
Most importantly this problem is not resolved by search and sampling methods
often employed for constructive methods, since they represent a very inefficient way of exploring the space of sequentially constructed solutions. 

\textbf{Contributions:} 
%\begin{itemize}
%	\item We propose a new type of sequential construction model utilizing a policy on the bounded joint action space of several vehicles which enables the creation of multiple tours in parallel.
%	\item We construct a more general and comprehensive state and action space to provide sufficient information about the status of other tours  based on an enhanced feature embedding for nodes, tours and vehicles.
%	\item Our model outperforms the state-of-the-art constructive model on the CVRP-TW while achieving comparable results on the existing CVRP benchmark. Moreover on two of three TW variants JAMPR creates solutions which are also significantly better than a comparable {(meta-)}heuristic approach specifically geared to routing problems.
%\end{itemize}
\begin{enumerate}
	%\item We tackle an important class of rich discrete optimization problems, the CVRP-TW, for the first time with direct machine learning (ML) methods.
	\item We show that existing constructive machine learning methods for other, less constrained discrete optimization 
	problems cannot easily be generalized to the CVRP-TW and hypothesize that this is due to consecutively constructing single tours.
	\item We propose a more expressive policy model JAMPR, that operates on several tours in parallel. To do so, we designed (i) a more  comprehensive state and action space than existing methods, providing sufficient information about other tours based on an enhanced feature embedding for nodes, tours and vehicles, and (ii) a policy decoder model that jointly selects both, the next location to visit and the tour to which this location should be added.
	\item In experiments on three variants of the CVRP-TW (\textit{hard}, \textit{partly-soft}, \textit{soft}) we show that our method provides vastly better solutions than existing %state-of-the-art 
	discrete optimization methods as well as state-of-the-art
	ML methods adapted to cope with time windows for two of these settings,
	while being on par with the discrete optimization methods for the third setting, still 
	outperforming the adapted machine learning methods there.
\end{enumerate}

%%%%%%%%%%%%%%%%%%%%%%%%%%%%%%%%%%%%%%%%%

\section{Related Work} \label{sec:related_work}

The first deep learning model for sequential solutions to VRPs was proposed by \citet{nazari_reinforcement_2018} who adapted the \textit{Pointer Network (PtrNet)} of \citet{vinyals_pointer_2015} to work on the CVRP. The original \textit{PtrNet} was based on a supervised learning approach later extended to reinforcement learning (RL) by \citet{bello2016neural}, both only for the TSP. \citet{nazari_reinforcement_2018} dropped the original RNN part of the encoder model completely and replaced it with a linear embedding layer with shared parameters. The decoder remained a RNN with pointer attention which sequentially produces a permutation of the input nodes.
The more recent model of \citet{kool2018attention} replaced this architecture with an adapted transformer model employing self-attention \citep{vaswani2017attention} instead. 
Concurrently \citet{deudon_learning_2018} combined the \textit{PtrNet} model with heuristic post-processing to tackle the TSP. 
This problem is also the focus of more recent works \citep{fu_targeted_2019, xing_graph_2019} which employ Monte-Carlo Tree Search similar to Alpha-Go \citep{silver2016mastering}. \citet{chen2019learning} propose an RL based improvement approach that iteratively chooses a region of a graph representation of the problem and then selects and applies established local heuristics. This approach was further improved by a disruption operator introduced by \citet{lu_learning_based_2019}. \citet{khalil_learning_2017} propose a Q-learning based method for the TSP and other graph related problems while \citet{kaempfer_learning_2019} use a supervised approach involving permutation invariant pooling to tackle the Multiple TSP. Another emerging approach by \citet{gasse2019exact} uses ML to find exact combinatorial solutions.

First machine learning approaches to tackle the CVRP-TW (with hard TW) have only been proposed very recently. To the best of our knowledge up to date there exist only two (improvement-based) methods. \citet{gao2020learn} use an enhanced version of the graph attention network \citep{velivckovic2017graph} to learn a heuristic for Very Large-scale Neighborhood Search \citep{shaw1998using} including improvement and destruction operators similar to \citep{lu_learning_based_2019}. While they are able to tackle large instances of up to 400 nodes, the performance gains compared to standard heuristic selection methods are only around 4-5\%. \citet{silva2019reinforcement} learn the sequence of 8 different neighborhood functions for a Variable Neighborhood Descent heuristic \citep{mladenovic1997variable} with tabular Q-learning. 
However, they only learn on a per-instance basis re-initializing the Q-table with zeros for each new instance. 
Instead we focus on the general machine learning approach to learn a policy
based on the whole distribution of problem instances and build our model on top
of the most advanced constructive method, the attention model \citep{kool2018attention}.
A recent survey on combinatorial optimization with machine learning techniques was presented by \citet{bengio_machine_2018} and \citet{guo2019solving} while more traditional heuristic approaches are discussed e.g.\ in \citep{toth2014vehicle}.

%Online Vehicle Routing With Neural Combinatorial Optimization and Deep Reinforcement Learning \cite{yu_online_2019}  RL, Struc2Vec + RNN \\

%%%%%%%%%%%%%%%%%%%%%%%%%%%%%%%%%%%%%%%%%

\section{Problem setting and base model} \label{sec:preliminaries}

%%%
\subsection{Vehicle routing problems}\label{sec:preliminaries:vrp}

\paragraph{CVRP}
%The simplest type of vehicle routing problem apart from the TSP is the (symmetric) capacitated Vehicle Routing Problem 
The Capacitated Vehicle Routing Problem can be defined on a graph $\mathcal{G} = \{\mathcal{V}, \mathcal{E}, q, c\}$ with node set $\mathcal{V}$ 
consisting of $N+1$ nodes, one depot node $0$ and $N$ customer nodes $1,...,N$. Each node $i$ has attributes including its Euclidean coordinates $r_i \in \mathbb R^2$ and a demand $q_i > 0,\ i \in \mathcal{V'}$ (where we define $\mathcal{V'} = \mathcal{V}$\textbackslash $\{0\}$) that needs to be satisfied. For the depot we set $q_0 = 0$.
The edge weights $c_{ij}$ of edges %$\mathcal{E} = \{ e_{ij} = e_{ji};\ i,j \in \mathcal{V},\ i \neq j \}$ 
$\mathcal{E} = \big\{ \{i,j\}\mid i,j\in \mathcal{V}, i\neq j \big\}$ are given by the transit costs
from node $i$ to node $j$. 
Furthermore there are $K$ homogeneous vehicles with same capacity $Q > 0$ representing the fleet. Without loss of generality we normalize demands $\tilde{q}_i = \frac{q_i}{Q}$ and set $\tilde{Q} = 1$. 
The tour $s_k$ of vehicle $k \in K$ is a sequence of indices w.r.t.\ a subset of all customers nodes %$v(s_k) \subseteq \mathcal{V'}$ 
representing the order in which vehicle $k$ visits the respective nodes. Furthermore every tour implicitly starts and ends at the depot. 
The set of sequences $S = \{s_1, ..., s_{K}\}$ constitutes a solution to the problem instance when 
(1) all customers are visited, i.e.\ 
%$v(S) = \mathcal{V'}$, 
$\bigcup_{s\in S} v(s) = \mathcal V'$,
(2) no customer is visited more than once, i.e.\ 
$v(s_k) \cap v(s_l) = \emptyset; \forall k,l \in K,\ k\neq l$
and 
(3) all tours respect the capacity constraint 
%$\lambda_q(s_k) := \sum_{i \in s_k} \tilde{q}_i \leq \tilde{Q}$.
$\sum_{i\in v(s_k)} q_i \le 1,\ \forall k \in K$.
%For the standard problem there exist two different objectives which is either the minimization of the total distance traveled by all vehicles or alternatively the minimization of the number of required vehicles (optionally with a limit on the total distance).

%%
\paragraph{CVRP-TW}
This extension of the CVRP is concerned with customers which can only be served within a specific time window. Moreover the service at each customer $i \in \mathcal{V'}$ requires a specific service duration $h_i$. 
A time window (TW) consists of a tuple $[a_i, b_i], a_i \leq b_i$ 
where $a_i$ and $b_i$ are upper and lower bounds on the possible arrival time.
Due times $\tau$ are included by setting $b_i = \tau - h_i$. The TW of the depot $[a_0, b_0]$ constitutes the available planning horizon 
regarding earliest possible departure from and latest return to depot.
Edge weights $c_{ij}$ represent transit costs in time units 
and without loss of generality include the service duration $h_i$ of the node $i$ from which the vehicle departs. 
A more comprehensive overview of the VRP and its variants can be found in \citep{toth2014vehicle} and \citep{braysy2005vehicle}.

%%%
\subsection{Learning-to-Optimize problem}\label{sec:preliminaries-ss:learning}
%According to the definition of the optimization problems above we are concerned with learning a stochastic policy $\pi(s \mid \text{\textbf{x}}, \theta)$ parameterized by learnable parameters $\theta$ which produces a solution $s$ for the corresponding problem instance $\text{\textbf{x}}$. In factorized form this can be written as:
%\begin{equation}\label{eq:pi}
%	%\pi(s \mid \text{\textbf{x}}, \theta) = \prod_{t=1}^{T} \pi(s_t \mid \text{\textbf{x}}, s_{1:t-1}, \theta)
%	\pi(s \mid \text{\textbf{x}}, \theta) = \prod_{t=1}^{T} \pi(s^{(t)} \mid \text{\textbf{x}}, s^{(1:t-1)}, \theta)
%\end{equation}
%where $T$ is the number of required decoding steps depending on $N$ and the number of used vehicles. 
%$\pi$ is represented by an encoder-decoder model which consecutively adds nodes to tours in solution $s$. %It is based on the architecture of \citet{kool2018attention} that is detailed in the next section.
While classical discrete optimization algorithms tackle each VRP from scratch and
in isolation, the Learning-to-Optimize problem consists of finding solutions to a problem, given, 
we have seen and solved many such problems (from the same distribution of problems) already,
without having access to the optimal solutions themselves. Given a sampler
for problems $(c, q, a, b) \sim p$ and a cost function $\operatorname{cost}$, learn a solver $\hat s$
that minimizes the cost of solutions for future problems from the same distribution
\begin{equation}\label{eq:l2o}
\min  \mathbb{E}_{(c,q,a,b)\sim p}(\operatorname{cost}(\hat s(c,q,a,b)))
\end{equation}

%%%
\subsection{Attention model for the sequential construction of VRP solutions}\label{sec:preliminaries-ss:base_model}
%The attention model (AM) \citep{kool2018attention} is an encoder-decoder architecture employing self-attention \citep{vaswani2017attention}. 

Constructive methods \citep{nazari_reinforcement_2018, kool2018attention} treat the Learning-to-Optimize problem as a sequential decision
making problem which constructs the solution piece by piece. The most advanced model of this class,
the attention model (AM) \citep{kool2018attention}, constructs
one route $s$ at a time by treating all not yet visited locations $i$ as actions. It learns a policy model
$
\pi( i^{(t)} \mid c,q,a,b, s^{(t-1)}; \theta)
$ and consequently $s^{(t)} = [ s^{(t-1)}; i^{(t)} ]$.

\paragraph{Encoder}
The encoder model takes the node-wise features $x_i = (r_i, q_i) \in \mathbb R^3,\ i \in \mathcal{V}$ consisting of coordinates $r_i$ and demand $q_i$ as input and first applies a linear projection $z^{0}_i = W^{\text{in}} x_i + b^{\text{in}}$ to create an initial embedding $z^{0}_i \in \mathbb R^{d_{\text{emb}}}$ with embedding dimension $d_{\text{emb}}$. 
The main part of the encoder consists of a stack of three self-attention blocks (SA) producing embeddings 
\begin{equation}
\omega^{\text{node}}_i = \SA(\ \SA(\ \SA(z^{0}_i, Z^{0}))\ )),\ i \in \mathcal{V}
\end{equation}
where $Z^{0} = (z^{0}_0,\ldots,z^{0}_N)$ is the sequence of initial embeddings.
Each block consists of a multi-head attention layer (MHA) \citep{vaswani2017attention},
an element-wise fully connected layer (FF), each in turn then followed by a residual connection (res) \citep{he2016deep} and a batch normalization layer (BN) \citep{ioffe2015batch}:
\begin{equation}\label{eq:}
z^{l}_i = \SA(z^{l-1}_i, Z^{l-1}) = \BN(\ \FFres(\ \BN(\ \MHAres(z^{l-1}_i, (z^{l-1}_0,\ldots,z^{l-1}_N))\ )))
\end{equation}

The multi-head attention layer is a linear combination of $H=8$ single-head attention (SHA) layers
each taking a slice of all the input elements (see appendix A for more details):
\begin{align}
\MHA(z, Z; W) & = \sum_{h=1}^H\ W^{\text{head}}_h  
\operatorname{SHA}(z_{\operatorname{slice}(h)}, Z_{.,\operatorname{slice}(h)};
%W^{\text{query}}_h, W^{\text{key}}_h, W^{\text{value}}_h).
W)
\end{align}
A single-head attention layer is a convex combination of  linearly transformed elements
with parametrized pairwise attention weights:
\begin{align}
\operatorname{SHA}(z, Z; W) & = \sum_{j=1}^{|Z|}
\operatorname{attn}( z, Z; W^{\text{query}}, W^{\text{key}})_j \, W^{\text{value}} Z_j
\\ \operatorname{attn}(z, Z; W^{\text{query}}, W^{\text{key}})
& =  \operatorname{softmax} \left(  \frac{1}{\sqrt{d_{\text{key}}}} z^T (W^{\text{query}})^T W^{\text{key}} Z_j  
\mid_{ \ j=1,\ldots,|Z|} \right), \quad z\in Z
\end{align}
Fully connected and residual layers are defined as usual:
\begin{align}
\MHAres(z_i, Z; W) & = z_i + \MHA(z_i, Z; W)             
\\ \FF(z_i; W, b) & = \max (0, W z_i + b)
\\ \FFres(z_i; W, b) & = z_i + \FF(z_i; W, b).
\end{align}

\paragraph{Decoder}
%The original decoder model of \citet{kool2018attention} does sequential decoding consecutively producing tours one node at a time. 
At decoding step $t \in \{1, ..., T\}$ the decoder takes a context $C^{(t)}$ and selects the next node to add to the current tour by attending over the sequence ${M = (\omega^{\text{node}}_0, \omega^{\text{node}}_1, ..., \omega^{\text{node}}_N)}$. The context $C^{(t)}$ consists of the concatenation of the graph embedding $\omega^{\text{graph}} = \frac{1}{N+1}\sum^N_{i=0} \omega^{\text{node}}_i$, the remaining capacity of the current vehicle $Q^{(t)}_f$ and the embedding $\omega^{\text{last}} = \omega^{\text{node}}_{\operatorname{last}(s)}$ of the preceding node (the depot in case of a new tour):
\begin{equation}\label{eq:context_org}
C^{(t)} = [\omega^{\text{graph}} ; Q^{(t)}_f ; \omega^{\text{last}}]
\end{equation}
where $[\ \ ;\ ]$ represents the concatenation operator. 
Its corresponding dimension is ${d_C = 2d_{\text{node}} + 1}$ with $d_{\text{node}}$ as the dimensions of the node embedding.
The decoder consists of a multi-head attention layer
and a subsequent layer with just attention weights operating on a masked input sequence,
where values of alternatives, which are ruled out by hard constraints of the routing problems, are set to $-\infty$ and thus yield weights of zero:
\begin{equation}% \label{eq:context_glimpse}
p_i = \pi(s_t = i \mid x, s_{1:t-1}, \theta) = \operatorname{attn}(  \MHA(C^{(t)}, M), \operatorname{mask}(M))
\end{equation}

%%%%%%%%%%%%%%%%%%%%%%%%%%%%%%%%%%%%%%%%%

\section{Attending multiple routes (JAMPR)} \label{sec:approach}

% careful when defining costs, since they should be negative. Therefore the sign of the penalty needs to be correct as well

Regarding the CVRP-TW, existing sequential construction approaches perform quite poorly because of the independent sequential construction of single tours and the very limited information on which each decision of the agent is based. We address these shortcomings in our Joint Attention Model for Parallel Route-Construction: JAMPR (pronounce "jumper") which introduces two major extensions to AM (described in section \ref{sec:preliminaries-ss:base_model}).
%our model by several adaptions and extensions of the standard Attention Model (AM) described in section \ref{sec:preliminaries-ss:base_model}. 

%\subsection{Defining suitable state and action spaces}
While the sequential construction of single tours is very efficient in terms of memory and computational complexity and was shown to work quite well for the CVRP, it faces some major challenges for more complex routing settings. 
%In general in deep RL the state space of the respective problems is very large and therefore is normally approximated by a parameterized function represented by a NN. %An important assumption is that the state embedding provides sufficient information about the problem state at any given time (XXX). 
Existing approaches appear to be sub-optimal since at time step $t$ in the creation of a specific tour the internal state representation for the decoder given by context $C^{(t)}$ and $M$ %, which is fed to the decoder in place of the sequence $Z$ 
comprises very limited information about the other existing tours. 
%The context consists of 
%$\omega^{\text{graph}}$, remaining capacity of current vehicle $Q_f$ and the embedding $\omega^{\text{pre}}$ of the preceding node. 
%$M$ is just the node embeddings $\omega^{\text{node}}$. 
Obviously in sequential decoding for the first tour there is no information about later tours. However, even for later tours the context only provides information about the preceding node. Nodes that are not available since they were already served by other tours are only considered by the masking scheme when calculating the compatibilities in the decoder, %$u_{i}$ and $\tilde{u}_{i}$
effectively pruning the action space but not providing any additional information.
%A first naive approach would aim to provide information about all existing tours in the context, e.g.\ via an embedding of a plan including all currently existing tours. However this only solves part of the problem, since in sequential construction early tours still have no information about later tours.
%To alleviate this problem we propose our Joint Attention Model for Parallel Route-Construction: JAMPR (pronounce "jumper") which introduces two major extensions to AM. 

\begin{figure}[h!t]
	\centering
	\includegraphics[width=1.0\textwidth]{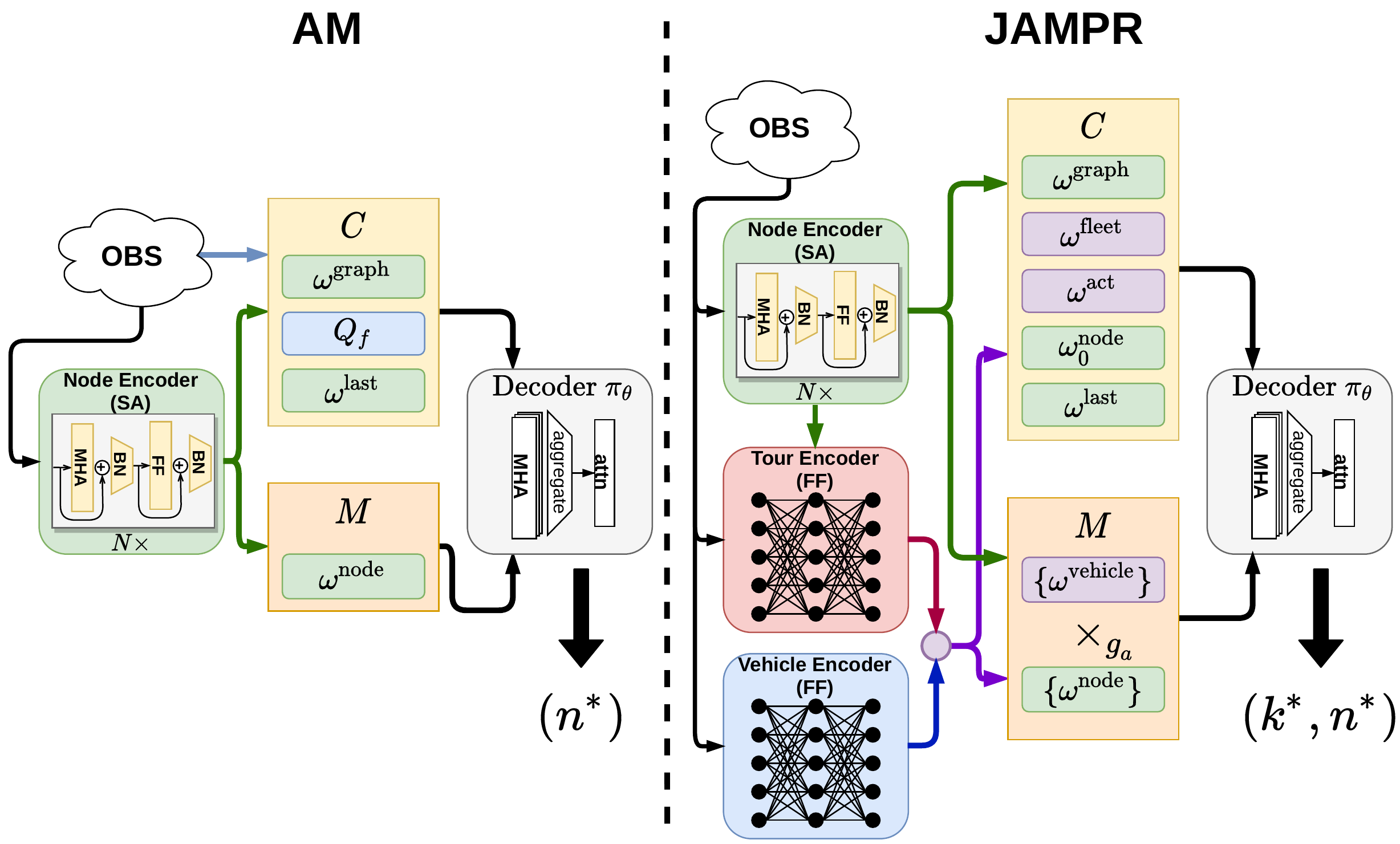}
	\caption{
		Illustration of the model architectures of JAMPR and AM \citep{kool2018attention} for one decoding step. Our model employs 2 additional encoders producing a comprehensive context $C$ and an enhanced joint action space represented by $\bar{M}$. %OBS indicates the state observations of the problem instance at the current step.
		OBS are the state observations of the problem at the current step.
	}
	\label{fig:architectures}
\end{figure}

\subsection{Comprehensive context information}\label{sec:approach-ss:context}
In order to provide sufficient information to the decoder at any time step we create a more comprehensive context by extending the state representation and adding two additional encoders for tours and vehicles. To simplify notation the whole section omits the index $(t)$ of the current time step. 

First we consider a state representation including a \textit{tour plan} $S$, represented by a $K \times L$ matrix consisting of $K$ tours of maximal length $L$ which is initialized with all zeros. When a node is added to a tour $k$ we insert the corresponding index in place of the first zero, the second node index in place of the next zero and so on.

The %\textit{vehicle encoder} 
additional encoders 
encode (i) \textit{vehicle features} $v_k$, consisting of
the vehicle index $k$,
the current return cost to the depot,      %  distance to the depot (in case of TW the corresponding return cost)
the current position of the vehicle (given by the coordinates of the last served node),
and the current time of each vehicle,
as well as (ii) the tour itself, consisting of the nodes visted so far.
The vehicle features are encoded by
a feed forward NN $g_v$,    % with shared parameters
the tour by the average output of a second feed forward NN $g_s$ fed with the embeddings $\omega^{\text{node}}_i$
of the visited nodes $i$ so far:
\begin{align*}
\omega^{\text{vehicle}}_k = [ g_v(v_k); \frac{1}{L}\sum_{i \in s_k} g_s(\omega^{\text{node}}_i) ]
\end{align*}

Our \textit{context} also includes information about the whole fleet
and the current active vehicles (see section \ref{sec:approach-ss:action_space}),
simply as the averages of their embeddings:
\begin{align*}
\omega^{\text{fleet}} = \frac{1}{K}\sum^{K}_{k = 1}\omega^{\text{vehicle}}_k, \quad 
\omega^{\text{act}} =  \frac{1}{K_{\text{act}}}\sum_{k \in K_{\text{act}}}\omega^{\text{vehicle}}_k, \quad 
\omega^{\text{last}} = \frac{1}{K}\sum^{K}_{k=1} \omega^{\text{node}}_{\operatorname{last}(s_k)}
\end{align*}
where $\operatorname{last}(s_k)$ denotes the last visited node in tour $s_k$. Together with
the initial graph embedding $\omega^{\text{graph}}$ and
the embedding $\omega^{\text{node}}_0$ of the depot node,
the context embedding then is defined as
\begin{equation}\label{eq:context}
C = [\omega^{\text{graph}} ; \omega^{\text{fleet}} ; \omega^{\text{act}} ;  \omega^{\text{node}}_0 ; \omega^{\text{last}}].
\end{equation}
with a dimension of $d_C = 3d_{\text{node}} + 2d_{\text{vehicle}}$ where $d_{\text{node}}$ and $d_{\text{vehicle}}$ are the dimensions of the node and vehicle embedding respectively. The complete architecture of our model compared to the original AM is shown in figure \ref{fig:architectures}.

\subsection{Enhanced joint action space for parallel route construction}\label{sec:approach-ss:action_space} 
For highly constrained problems, we furthermore propose to construct several routes $S$ in parallel, and treat
all pairs of routes $s_k \in S$ and not yet visited locations $i$ as actions, i.e., to learn a policy model
$
\pi( k^{(t)},  i^{(t)} \mid c,q,a,b, S^{(t-1)}; \theta)
$. Then $S^{(t)}_{k^{(t)}} = [ S^{(t-1)}_{k^{(t)}}; i^{(t)} ]$ and all other $S^{(t)}_k = S^{(t-1)}_k$ ($k\neq k^{(t)}$)
stay the same.
%In order to relax the highly independent sequential construction of single tours we then define a joint action space enabling the construction of several routes in parallel. 
Therefore we combine the node embeddings $\omega^{\text{node}}$ with the vehicle embeddings $\omega^{\text{vehicle}}$ creating the joint combinatorial space of all nodes and vehicles which represents the new $M$: 
\begin{equation}\label{eq:M_prime}
M = \{\ g_a(\omega^{\text{vehicle}}_k, \omega^{\text{node}}_i) \mid k=1,\dots, K,  i \in \mathcal{V}\}
\end{equation}
with a NN $g_a: \IR^{d_{\text{node}} + d_{\text{vehicle}}} \rightarrow \IR^{d_M}$ that models the compatibility or affinity between a tour and a node according to
%$g_a(\omega^{\text{vehicle}}_k, \omega^{\text{node}}_i)= [ \omega^{\text{vehicle}}_k \odot \omega^{\text{node}}_i ;
%(\omega^{\text{vehicle}}_k)^T \omega^{\text{node}}_i ]$, the concatenation of the elementwise ($\odot$) and the dot product.
\begin{equation}
g_a(\omega^{\text{vehicle}}_k, \omega^{\text{node}}_i)= 
\big( 
W_1\omega^{\text{node}}_i + 
W_2\omega^{\text{vehicle}}_k +
W_3[\omega^{\text{vehicle}}_k \odot \omega^{\text{node}}_i ; (\omega^{\text{vehicle}}_k)^T \omega^{\text{node}}_i]
\big)
\end{equation}
the sum of the projected node and vehicle embedding and the projected concatenation of the elementwise ($\odot$) and dot products with suitable weight vectors $W$ to project the respective components to the shared dimension $d_M$.

Since the size of M is subject to combinatorial explosion w.r.t.\ $N$ and $K$ we fix the number of active vehicles for which decisions can be done concurrently to $m_{con}$ (at any time there are exactly $m_{con}$ vehicles active). If a vehicle has no feasible nodes left (because of capacity or time constraints) or it selects the depot node, it returns to the depot and is set inactive. Then a new tour is initialized by activating the next available vehicle. We define $M^{(t)}$ which changes according to the currently active vehicles $K^{(t)}_{\text{act}}$ at time step $t$ but has a constant size $|M^{(t)}| = m_{con} \cdot N+1$ as:
\begin{equation}%\label{eq:M_bar}
M^{(t)} = \{  g_a(\omega^{\text{vehicle}}_k, \omega^{\text{node}}_i) \mid k \in K^{(t)}_{\text{act}}, i \in \mathcal{V}\}
\end{equation}

\section{Experiments} \label{sec:experiments}
%%%

\subsection{Experiment setup}

\paragraph{Problem Details}
We evaluate JAMPR for three different variants of the CVRP-TW, which differ in the way they handle the respective time windows (TW) and consequently are concerned with different application scenarios: %\citep{fu2008unified}:

\begin{enumerate}
	\item \textbf{TW1} The first variant is the most well-known problem with \textit{hard} TW. Vehicles can only serve customers within the TW given by $[a_i, b_i]$. When too early they wait at no additional cost until %the earliest start time 
	$a_i$ while too late vehicles are not able to serve the respective customer.
	\item \textbf{TW2} For the second type of problem the latest arrival time $b_i$ is a soft constraint which can be violated by paying a corresponding penalty. The late deviation of vehicle $k$ is defined as 
	$\delta^k_{b_i} = \max(T_{ik} - b_i, 0)$
	where $T_{ik}$ is the arrival time of vehicle $k$ at node $i$. Vehicles that are too early wait at no cost until $a_i$. % like in variant 1.
	\item \textbf{TW3} The third variant involves soft constraints for both the upper and the lower bound of the TW. They can be violated by paying the respective penalty. The early deviation of vehicle $k$ is defined as 
	$\delta^k_{a_i} = \max(a_i - T_{ik}, 0)$.
\end{enumerate}

%The two later variants correspond to the problems of type \textit{1)} and \textit{2)} defined in \citet{fu2008unified}. 
Respective penalties involve a suitable penalty function $\lambda$ (e.g.\ $\lambda(x)= x$ or $\lambda(x)= x^2$) which is monotonously increasing in $\IR^{+}$.
Finally the cost function is defined wr.t.\ the total time required by all vehicles and the incurred penalties according to:
\begin{equation}\label{eq:cost}
\operatorname{cost} = \sum^K_{k=1} \left(\sum_{(i,j) \in s_k \uplus\ 0} c_{ij} + \alpha \lambda(\delta^k_{a_i}) + \beta \lambda(\delta^k_{b_i}) \right)
\end{equation}
where $\alpha, \beta \geq 0$ are respective weights, $j$ is the index following $i$ in the corresponding sequence $s_k \uplus\ 0$ where the $0$ index is included at start and end to take care of departure and return to the depot.

\paragraph{Data Generation}
For the CVRP-TW we sample suitable problem instances from a distribution based on the statistics of the R201 instance of the well-known benchmark problem set of \citet{solomon1987algorithms}. Capacities are set to $Q^{20} = 500$ and $Q^{50} = 750$ %and $Q^{100} = 1000$. 
for problem size 20 and 50 respectively.
The full time horizon is $[a_0=0,\ b_0=1000]$ while the service durations $h_i$ are uniformly set to 10. More details can be found in the supplementary material.
For the generation of instances for the CVRP we follow the sampling procedure of \citep{kool2018attention} and use the same vehicle capacities of $Q^{20} = 30$ and $Q^{50} = 40$ %and $Q^{100} = 50$ 
as well as the same validation and test sets.

\paragraph{Training and Hyper-Parameters}
% weight initialization?
We train our model with the policy gradient based on the established REINFORCE algorithm \citep{williams1992simple} with the rollout baseline proposed in \citep{kool2018attention}. We employ the Adam optimizer \citep{kingma2014adam} with a smooth learning rate decay schedule according to $\eta_{t} = (\frac{1}{1 + \gamma t}) \eta_{t-1}$ at epoch $t$ with decay factor $\gamma = 0.001$ and an initial learning rate of $10^{-4}$. Our node encoder consists of 3 SA blocks with dim=128, the tour and vehicle encoder both have a hidden dimension of 64. The tour encoder has 2 layers while the vehicle encoder uses 1 layer for the CVRP and 3 layers for the CVRP-TW. The embedding dimensions for all problems are given by $d_{\text{node}}=d_{\text{vehicle}}=128$. For the decoder we use a hidden dimension of 256. Furthermore we clip the norm of the gradients to 1. The model specific hyper-parameter $m_{con}$ controlling the number of concurrently constructed tours is tuned with a grid search between 1 and 4 on the validation set. We train the models for 50 epochs, each epoch involving 1,024,000 training instances with a problem size dependent batch size of $BS^{20}=512$ and $BS^{50}=128$. We find that our models are mostly converged after 50 epochs, but train them nevertheless for 100 epochs on the CVRP to be comparable to related work.
Our code will be made available with the published paper on github: \href{https://github.com/AnonymousAuthor}{https://github.com/AnonymousAuthors}.

\paragraph{Baseline models}
We compare our model to the current learning based state-of-the-art model for sequential route construction AM \citep{kool2018attention} which we adapt for the CVRP-TW by including the current time of the vehicle in the context and an updated masking scheme considering TW (AM\textsuperscript{+TW}).  
Furthermore we compare against the Google OR-Tools (GORT) library \citep{ortools} which frequently serves as established (meta-)heuristic baseline in the related work. We apply it with two different configurations regarding the underlying local search heuristic, for which we select either \textit{automatic selection (AU)} or \textit{guided local search (GLS)}.
%We apply it with several different configurations regarding the underlying local search heuristic, for which we select either \textit{automatic selection (AU)}, \textit{guided local search (GLS)}, \textit{simulated annealing (SA)} or \textit{tabu search (TS)}, where the three later methods required the specification of a suitable time limit.
Additionally we compare against related work for the standard CVRP including the VRP extension of the \textit{PointerNet (PtrNet)} \citep{nazari_reinforcement_2018},
the two improvement methods \textit{NeuRewriter} \citep{chen2019learning} and 
\textit{Learning to improve (L2I)} \citep{lu_learning_based_2019}, % and an extension of the Lin-Kernighan-Helsgaun solver 
and \textit{LKH3} \citep{helsgaun2017extension} %as well as 
and GORT as heuristic approaches (from \citep{lu_learning_based_2019}).

\paragraph{Inference Setup}
We report results for greedy decoding and sampling where 1280 solutions are sampled from the stochastic policy $\pi$ (as one batch) and we report the best one. For AM\textsuperscript{+TW} we compare results for the same number of samples (1280) and for approximately the same inference time assuming sequential processing of sample batches (10240 samples). The reported inference times are the approx. time required to solve a single instance (BS=1) averaged over the test set. 
Since run times are hard to compare because of different batch size and parallelization capability we argue for this practical approach focusing on the most common use case where one single problem instance is solved at a time. However, the learned models are much better in leveraging parallelization via mini-batch processing on a GPU to quickly solve larger amounts of instances. Consequently because of the varying evaluation protocols in the related work most inference times are not comparable and the respective number $k$ of used vehicles is not known (marked by "-"). We learn our models on a single GPU (Nvidia 1080TI) while GORT is run on CPU (Intel Xeon E5-2670v2). %(Intel Core i7-7700K).

%%%
\subsection{Results}
Table \ref{tab:results} shows the results of our experiments on the three described variants of the CVRP-TW. We report the average cost on the test set (10000 instances). %according to Eq. \ref{eq:cost}. %The best result per problem is marked in \textbf{bold}. 
%Because of limited resources and time constraints before the deadline the models for instances of size 50 could only be trained for 50 epochs. While the AM\textsuperscript{+TW} baseline had already fully converged, JAMPR should still benefit from further training. We will report fully converged results with the camera ready version of the paper. 
According to \citep{braysy2005vehicle} the main objective in most CVRP-TW is to minimize the number of required vehicles to serve all nodes while respecting all constraints. The minimization of distance or time is usually only a secondary objective. 
For VRP these objectives have been shown to be conflicting so that the reduction in the number of used vehicles $k$ often causes an increase in total distance \citep{braysy2001local}. For this reason we report the average $k$ as well. 

\begin{table}[h!t]
	\caption{Comparison of results for CVRP and the CVRP-TW variants. Best results for each problem are \textbf{bold}. Results taken from related work are marked with \textdagger.}
	\label{tab:results}
	\centering
	%\footnotesize
	\small
	\begin{tabular}{llrrrrrr} %|ll}
		\toprule
		& \textbf{Model} & \multicolumn{3}{c}{\textbf{N=20}} & \multicolumn{3}{c}{\textbf{N=50}} \\%& \multicolumn{2}{c}{\textbf{N=100}} \\
		& & \texttt{cost} & \texttt{k} & \texttt{t\textsubscript{inf}} & \texttt{cost} & \texttt{k} & \texttt{t\textsubscript{inf}} \\ 
		\hline
		\hline
		\multirow{7}{*}{\rotatebox[origin=c]{90}{\parbox[c]{1cm}{\centering \textbf{TW1}}}} % hard
		& GORT - \textit{AU} 				& 2577.08  & 4.18 & 0.22s & 4344.88  & 6.30 & 1.76s \\
		& GORT - \textit{GLS} 				& 2522.85  & 4.11 & 8.00s & 4213.23  & 6.15 & 8.06s \\
		& AM\textsuperscript{+TW} (greedy) 	& 3766.88  & 5.29 & 0.05s & 7189.43  & 9.42 	& 0.12s \\
		& AM\textsuperscript{+TW} (sampl.) 	& 3041.24  & 5.74 & 0.12s & 7327.09  & 11.92 & 0.38s \\
		& AM\textsuperscript{+TW} ($t_{10240}$) & 2750.06  & 5.27 & 0.95s & 6878.84  & 11.28 & 3.08s \\
		& JAMPR (greedy) 	 				& 1862.40  & 2.25 & 0.10s & 3055.94  & 5.42 & 0.24s	\\
		& JAMPR (sampl.) 					& \textbf{1716.60}  & 2.29 & 0.86s & \textbf{2691.55}  & 4.03 & 3.07s \\
		\hline
		\multirow{7}{*}{\rotatebox[origin=c]{90}{\parbox[c]{1cm}{\centering \textbf{TW2}}}} % wait
		& GORT - \textit{AU} 				& 635.06  & 4.23 & 0.37s & 1123.82  & 6.72 & 3.81s 	\\
		& GORT - \textit{GLS} 				& \textbf{619.57}  & 4.14 & 8.30s & \textit{\textbf{1119.07}}  & 6.67 & 8.09s \\
		& AM\textsuperscript{+TW} (greedy)	& 7615.69  & 2.00 & 0.05s & 40245.40  & 2.00 	& 0.12s \\
		& AM\textsuperscript{+TW} (sampl.) 	& 1572.31  & 6.56 & 0.11s & 7712.35  & 9.34 	& 0.34s \\
		& AM\textsuperscript{+TW} ($t_{10240}$) & 1387.30  & 6.46 & 0.90s & 6730.55  & 9.99 & 2.70s \\
		& JAMPR (greedy) 	 				& 674.72  & 4.32 & 0.11s & 1273.20  & 6.02 & 0.25s 		\\
		& JAMPR (sampl.) 					& \textit{\textbf{620.68}}  & 4.19 & 0.92s & \textbf{1116.76}  & 5.64 & 2.32s 	\\
		\hline
		\multirow{7}{*}{\rotatebox[origin=c]{90}{\parbox[c]{1cm}{\centering \textbf{TW3}}}} % soft
		& GORT - \textit{AU} 				& 1317.81  & 4.07 	& 1.07s & 2707.72  & 6.12 & 20.63s 	\\
		& GORT - \textit{GLS} 				& 1312.71  & 4.11 	& 8.00s & 2753.66  & 6.18 & 8.02s 	\\
		& AM\textsuperscript{+TW} (greedy) 	& 3101.21  & 2.00 & 0.05s & 26467.34  & 2.00 & 0.12s 	\\
		& AM\textsuperscript{+TW} (sampl.) 	& 1412.16  & 3.42 & 0.12s & 4161.24  & 7.15 & 0.34s 	\\
		& AM\textsuperscript{+TW} ($t_{10240}$) & 1318.56  & 3.23 & 0.88s & 3941.47  & 6.77 & 2.67s  \\
		& JAMPR (greedy) 	 				& 1002.81  & 1.00 & 0.10s & 3158.26  & 2.01 & 0.23s  \\
		& JAMPR (sampl.) 					& \textbf{844.35}  & 1.39 & 0.78s & \textbf{1947.65}  & 2.29 & 2.13s 	\\
		\hline
		\hline
		\multirow{10}{*}{\rotatebox[origin=c]{90}{
				\parbox[c]{4cm}{\centering \textbf{CVRP} \\ \footnotesize{w/o time windows \\ (for comparison)}}
			}} 
		& LKH3 \textdagger 				& 6.14 & - & - & 10.38 & - & - 				\\
		& GORT \textdagger 				& 6.43 & - & - & 11.31 & - & - 				\\
		& NeuRewriter \textdagger  		& 6.16 & - & - & 10.51 & - & - 				\\
		& L2I \textdagger 				& \textbf{6.12} & - & - & \textbf{10.35} & - & - \\
		& PtrNet \textdagger (greedy) 	& 6.59 & - 		& - 		& 11.39 & - 	& - \\
		& PtrNet \textdagger (beam) 	& 6.40 & - 		& - 		& 11.15 & - 	& - \\
		\cmidrule(l{5pt}llr{5pt}){2-8}
		& AM (greedy) 					& 6.40  & 5.90 	& 0.03s 	& 10.98  & 7.00		& 0.08s 	\\
		& AM (sampl.) 					& 6.25  & 5.22 	& 0.05s 	& 10.62  & 7.65 	& 0.20s 	\\
		& JAMPR (greedy) 				& 6.47  & 4.06 	& 0.11s 	& 11.44  & 7.31 	& 0.23s 	\\
		& JAMPR (sampl.) 				& 6.26  & 3.97 	& 0.84s 	& 10.84  & 7.13 	& 2.81s 	\\
		\bottomrule
	\end{tabular}
	
\end{table}

JAMPR outperforms AM\textsuperscript{+TW} on all CVRP-TW by a large margin. Compared to GORT we also achieve significantly better results on TW1 and TW3. For TW2 our model finds solutions of similar quality while being much faster overall. Although inference for AM\textsuperscript{+TW} for greedy decoding and sampling is faster, the results are much worse. Taking the speed into consideration we perform inference for  AM\textsuperscript{+TW} with 8 times as many samples ($t_{10240}$). While this improves the results, it is still significantly worse than JAMPR for all CVRP-TW and for GORT for all but TW3. For the standard CVRP our model performs slightly worse than AM but still achieves reasonable results showing that the model is also useful for the vanilla CVRP without TW. 
Although it is no explicit part of our model or objective function, we find that our model generally creates solutions with a much lower average $k$ than AM\textsuperscript{+TW} as well as GORT.  

%An additional ablation study regarding the max concurrency controlled by $m_{con}$ can be found in the supplementary material.
In general it seems that for bigger problem instances requiring a larger total number of vehicles, a higher concurrency controlled by $m_{con}$ is advantageous. In contrast for small instances that only require 2-3 vehicles, $m_{con}=1$ or $m_{con}=2$ works better. Furthermore larger $m_{con}$ seems to be helpful for more constrained problems like TW1 with hard time windows while $m_{con}=1$ is sufficient for the TW3 where the only hard constraint is the capacity. As expected AM\textsuperscript{+TW} performs quite poorly on all TW problems because of the limited information in the context and the simplistic action space. While large numbers of samples help to some extend to improve the AM\textsuperscript{+TW} results for small problems ($t_{10240}$), this effect seems to be diminished by the combinatorial explosion of the solution space for larger instances.  
%Another interesting finding is that AM\textsuperscript{+TW} does not learn to use more than 2 tours in greedy decoding for CVRP-TW2 and CVRP-TW3. While it is possible to serve all customers with only 2 vehicles, the reduction of late arrivals for more vehicles is significant as shown by GORT and JAMPR.

%%%%%%%%%%%%%%%%%%%%%%%%%%%%%%%%%%%%%%%%%

\section{Conclusion} \label{sec:conclusion}
In this paper we introduced a new constructive model for solving highly constrained VRP.  
It is the first attempt for a learned constructive approach to tackle the CVRP-TW where our model shows strong results even outperforming a problem specific meta-heuristic approach. This contributes to the advancement of discrete optimization approaches based on learned models which this work successfully extends to much more difficult problem settings. 
In the future we plan to take a look at online optimization scenarios. In that case the comprehensive context should enable online decoding or respectively partial route construction, since the provided comprehensive information about the problem state transforms the decoding procedure into an approximate Markov Decision Process respecting the Markov property.

\bibliography{references}

% APPENDIX
%%%%%%%%%%%%%%%%%%%%
\renewcommand\thesection{\Alph{section}}
\renewcommand\thesubsection{\thesection.\Alph{subsection}}
\setcounter{section}{0}
\renewcommand*{\theHsection}{chX.\the\value{section}}
%%%%%%%%%%%%%%%%%%%%%%%%%%%%%%%%%%%%%%%%%

%\newpage
%\section{Supplementary Material} \label{sec:appendix}

\section{Attention Mechanism}

A self-attention layer \citep{vaswani2017attention} transforms an input sequence $Z:= (Z_1, Z_2, \ldots, Z_J)$ of vectors $Z_j\in\mathbb{R}^{d_{\text{in}}}$
into an output sequence of vectors in $\mathbb{R}^{d_{\text{out}}}$ allowing to model dependencies between
all input elements based on their values, but not on their position.
A \textbf{single-head attention layer (SHA)} is a convex combination of  linearly transformed elements (called \textbf{values}):
\begin{equation}
\operatorname{SHA}(z, Z; W^{\text{query}}, W^{\text{key}}, W^{\text{value}}) = \sum_{j=1}^{|Z|}
\operatorname{attn}( z, Z; W^{\text{query}}, W^{\text{key}})_j \, W^{\text{value}} Z_j
\end{equation}

with parametrized pairwise \textbf{attention weights}
\begin{align}
\operatorname{attn}(z, Z; W^{\text{query}}, W^{\text{key}})
& =  \operatorname{softmax}(  \frac{1}{\sqrt{d_{\text{key}}}} z^T (W^{\text{query}})^T W^{\text{key}} Z_j  
\mid_{ \ j=1,\ldots,|Z|} ), \quad z\in Z 
\\ & \quad W^{\text{value}}\in\mathbb{R}^{d_{\text{out}}\times d_{\text{in}}},
W^{\text{query}}, W^{\text{key}} \in\mathbb{R}^{d_{\text{key}}\times d_{\text{in}}}, d_{\text{key}}\in\mathbb{N}
\end{align}
$W^{\text{query}}z$ is usually called the \textbf{query}, $W^{\text{key}} Z_j$ the \textbf{key} for element $j$.

A \textbf{multi-head attention layer (MHA)} is a linear combination of several single-head attention layers
(with ouput size $d_{\text{value}}$)
each taking a slice of all the input elements:
\begin{align}
\MHA(z, Z; W) & = \sum_{h=1}^H W^{\text{head}}_h  
\operatorname{SHA}(z_{\operatorname{slice}(h)}, Z_{.,\operatorname{slice}(h)};
W^{\text{query}}_h, W^{\text{key}}_h, W^{\text{value}}_h)
\\ \operatorname{slice}(h; H,d_{\text{in}}) & = ( 1+h\Delta h, 2+h\Delta h, \ldots, \Delta h + h\Delta h ), \quad \Delta h:= \frac{d_{\text{in}}}{H}
\\ & \quad W = (W^{\text{head}}_h, W^{\text{query}}_h, W^{\text{key}}_h, W^{\text{value}}_h)_{h=1:H}
\\ &  \quad W^{\text{head}}_h\in\mathbb{R}^{d_{\text{out}  }\times d_{\text{value}}},
W^{\text{value}}_h\in\mathbb{R}^{d_{\text{value}  }\times d_{\text{in}}}, d_{\text{value}}\in\mathbb{N},
\end{align}

\section{Batch Normalization}
The batch normalization component (BN) \citep{ioffe2015batch} used in the encoder can be defined as
\begin{equation}\label{eq:BN}
	\BN(z_i; W, b) = W \odot \frac{z_i - \mu_{\mathcal{B}}}{\sqrt{\sigma^2_{\mathcal{B}} + \epsilon}} + b
\end{equation}
with mean $\mu_{\mathcal{B}}$ and variance $\sigma^2_{\mathcal{B}}$ of mini-batch $\mathcal{B}$, noise term $\epsilon$ and learnable parameters $W$ and $b$.

\section{Recovering decisions from joint action space embedding}
To recover the original decision space regarding the selected vehicle $k^*$ and selected node $n^*$ from $M^{(t)}$ we use an index mapping $\varphi: \IN \rightarrow \IN^2$ mapping the index of each embedding vector back to the corresponding decision pair $(k^*, n^*)$:
\begin{equation}
\varphi (m) =  (k^*, n^*)
\end{equation}
where $m \in |M|$ is the selected index from the combinatorial space of $M^{(t)}$.

\section{Considerations regarding computational efficiency}
To make the proposed model computationally feasible we separate the embeddings into their respective static and dynamic components.
The routing problems we are concerned with in this paper are deterministic and static and correspondingly the embeddings $\omega^{\text{node}}$ and accordingly $\omega^{\text{graph}}$ are completely static and can be easily pre-computed. 

Furthermore we propose to construct several routes $S$ in parallel, learning a policy model
${
\pi( k^{(t)},  i^{(t)} \mid c,q,a,b, S^{(t-1)}; \theta)
}$
 where ${S^{(t)}_{k^{(t)}} = [ S^{(t-1)}_{k^{(t)}}; i^{(t)} ]}$ and all other $S^{(t)}_k = S^{(t-1)}_k$ ($k\neq k^{(t)}$)
stay the same.
Consequently 
$\omega^{\text{vehicle}}$ only changes in the dimension of one specific vehicle $k$ at each decoding step while it stays the same for all other vehicles. Therefore we only need to update the latent embedding of the currently selected vehicle $k$ with the encoder models and then just recompute the means for $\omega^{\text{fleet}}$ and $\omega^{\text{act}}$. The same argument holds for the static and dynamic components of $M$ where $\omega^{\text{node}}$ is static and can be pre-computed while $\omega^{\text{vehicle}}$ as well as the interactions $\omega^{\text{had}}$ and $\omega^{\text{dot}}$ need to be updated respectively.

%
%To improve the computational efficiency we absorb the weights $W_1$, $W_2$ and $W_3$ from Eq. 14 into the weights of $W^{\text{query}}, W^{\text{key}}, W^{\text{value}}$.
%
%Moreover for the static components we can even pre-compute the full projections of the operator $\psi$ in the decoder model which then are just added to the projections of the dynamic components:
%\begin{align}\label{eq:comp_eff}
%\begin{split}
%\rho & = W^{\text{stat}}_{\rho} C^{\text{stat}} + W^{\text{dyn}}_{\rho} C^{\text{dyn}} \\
%& = W^{\text{stat}}_{\rho} [\omega^{\text{graph}} ; \omega^{\text{node}}_0] + W^{\text{dyn}}_{\rho} [\omega^{\text{fleet}} ; \omega^{\text{act}} ; \omega^{\text{pre}}]
%\end{split} 
%\\[2ex]
%%\end{align}
%%\vspace{2cm}
%\begin{split}
%\kappa_i & = W^{\text{stat}}_{\kappa} m_i^{\text{stat}} + W^{\text{dyn}}_{\kappa} m_i^{\text{dyn}} \\
%& = W^{\text{stat}}_{\kappa} \omega^{\text{node}}_i + W^{\text{dyn}}_{\kappa} [\omega^{\text{vehicle}}_i ; \omega^{\text{had}}_i ; \omega^{\text{dot}}_i]
%\end{split} 
%\\[2ex]
%%\vspace{2cm}
%\begin{split}
%\nu_i & = W^{\text{stat}}_{\nu} m_i^{\text{stat}} + W^{\text{dyn}}_{\nu} m_i^{\text{dyn}} \\
%& = W^{\text{stat}}_{\nu} \omega^{\text{node}}_i + W^{\text{dyn}}_{\nu} [\omega^{\text{vehicle}}_i ; \omega^{\text{had}}_i ; \omega^{\text{dot}}_i]
%\end{split}
%\end{align}
%with $i \in |M|$ and respective projection weights $W$ for the concatenated latent feature vectors of the static and dynamic components..

\section{Reinforcement learning framework}
Similar to previous approaches we employ a policy gradient based on the REINFORCE algorithm \citep{williams1992simple} with baseline. 
We adopt the rollout baseline proposed by \citet{kool2018attention} since it was shown to work robustly well for this type of problem. Based on small preliminary experiments with a moving exponential average baseline and a learned critic model we can confirm these findings. 
For this reason we employ the aforementioned rollout baseline which corresponds to a greedy rollout of the best model checkpoint found so far. The evaluation of the model checkpoint is done on a separate validation set which is re-sampled each episode to prevent over-fitting. The significance of the performance difference is checked with a paired t-test with $\alpha=0.05$. We configure the baseline with the same specification as \citep{kool2018attention} including a warm-up of 1 epoch via exponential average ($\beta = 0.8$).

%Definition of reward w.r.t. objective?

\section{Data generation for the CVRP-TW}\label{sec:benchmark_data}
Here we give details for the generation of the data for the CVRP-TW training, validation and test set.
To implement a suitable data generation protocol we took a look on the well-known benchmark problem set of \citet{solomon1987algorithms} (\href{http://web.cba.neu.edu/~msolomon/problems.htm}{http://web.cba.neu.edu/\textasciitilde msolomon/problems.htm}). 
It consists of CVRP-TW instances representing different setups regarding 
\begin{enumerate}
	\item \textbf{Geographical Data:}\  There are three different approaches to sample the location of customers and the depot: uniformly at random \textbf{(R)}, clustered \textbf{(C)} and both random and clustered \textbf{(RC)}.
	\item \textbf{Scheduling Horizon:}\  The instances are sampled for two different scenarios, either a short horizon \textbf{(1)} for many vehicles serving small numbers of customers or a long horizon \textbf{(2)} where a smaller number of vehicles is able to serve a larger number of customers each.
	\item \textbf{TW width:}\  The width of the TW effects the flexibility of planning the tours within a specific time interval.
	\item \textbf{TW density:}\  The density of TW controls how many of the customers are constrained by hard TW, either 25, 50, 75 or 100\%.
\end{enumerate}

For our experiments we select the \textit{R201} instance of the benchmark as master sample since it suits our use case the best. It consists of randomly sampled geographical data with a long scheduling horizon and 100\% of the customers are constrained by short to medium sized TW. All service durations are 10 time units and the standard capacity of all vehicles for problem size 100 is $Q^{100} = 1000$. The service horizon is given by the depot time window $[a_0=0,\ b_0=1000]$. The corresponding demands $q_i$ of the instance have a mean of 17.24 and standard deviation of 9.4175. 

In order to sample similar instances we define the following constraints and sampling routines described below:
\begin{itemize}
	\item \textbf{Locations:}\\ The locations of the customers and the depot are sampled uniformly from the interval [0, 100].
	
	\item \textbf{Demands:}\\  For the demand $q_i$ of customer $i$ we sample from a normal distribution 
	$\hat{q} \sim \mathcal{N}(\mu, \sigma)$ with mean $\mu=15$ and standard deviation $\sigma=10$ and then clamp its absolute value to integer values between 1 and 42:\ $q = \min(42, \max(1, \lfloor|\ \hat{q}\ |\rfloor))$.
	
	\item \textbf{TW:}\\  In order to sample TW which are feasible regarding the travel time required from the depot to the corresponding customer, we 
	first define a sample horizon $h_i$ for each customer $i$ according to $h_i = [a_{\text{sample}}=\hat{h_i}, b_{\text{sample}}=b_0 - \hat{h_i}]$ where $\hat{h_i} = \lceil d_{0i} \rceil +1$ and $d_{0i}$ is the L2 distance from the depot to customer $i$. Then we sample the TW start time $a_i$ uniformly from $h_i$. The end time $b_i$ is calculated as 
	\begin{equation*}
		b_i = \min(\lfloor a_i + 300\epsilon \rfloor, b_{\text{sample}})
	\end{equation*}
	where $\epsilon$ is a noise term sampled from a standard normal distribution according to $\epsilon = max(|\ \hat{\epsilon}\ |, 1/100),\ \hat{\epsilon} \sim \mathcal{N}(0,1)$
 
\end{itemize}
Furthermore we use the same capacity $Q^{100} = 1000$. For smaller problem instances we set the respective capacities to $Q^{20} = 500$ and $Q^{50} = 750$. All service durations are 10 time units. Our validation and test sets as well as the generator code will be made available with the published paper.

\section{Additional details regarding the experiment setup}

\paragraph{Problems}
The problem specific weights for the penalty terms in the objective function we set specific to our use cases to: 
\begin{itemize}
	\item TW1: $\alpha=1.0,\ \beta=\infty$,
	\item TW2: $\alpha=0.0,\ \beta=0.5$,
	\item TW3: $\alpha=0.1,\ \beta=0.5$
\end{itemize}
with linear penalty functions $\lambda(x)= x$.

\paragraph{JAMPR}
For our model we allow a number of pre-mature returns for vehicles which do not use their full capacity in order to start new intermediate tours by activating vehicles. They are limited by the hyper-parameter  
$m_{pre}$ which we set to 3 for the CVRP and to 6 for the CVRP-TW. 

\paragraph{AM}
For the CVRP-TW we increase the hidden dimension of the decoder model from 128 to 256.

\paragraph{GORT}
Since GORT is only able to process integer valued data we scale all attributes of the standard problem instances by 100 and round to the next integer values. For that reason we also use the un-normalized original demands $q$ and capacity $Q$. The created solutions consist of the node indices in the corresponding tours and therefore can be evaluated in the same way as the learned models.
Furthermore the GORT \textit{GLS} model requires the specification of a suitable time limit for the local search procedure. We run the GORT \textit{GLS} baseline with a local search time limit of \textit{8s} per instance. If no feasible solution is found within this time limit, it is consecutively doubled until a feasible solutions is found.

\newpage
\section{Detailed experiment results}\label{sec:detailed_exp}
While working on the benchmark (see section \ref{sec:benchmark}) we experimented with an additional constraint for the GORT baseline, restricting waiting time to be less than an upper limit $\hat{\delta}^k_{a_i}$. After trying different limits, we found that a $\hat{\delta}^k_{a_i}$ of 10min works best. This improved the results on the benchmark w.r.t\ to our cost function by a large margin. Therefore we re-run the adapted GORT baseline (marked as \textit{GORT*}) for our large sampled test set as well. The results can be found in table \ref{tab:detailed_results}. While the adapted constraint improves the results in some cases by balancing the objectives of minimizing total distance and waiting times, the general finding is still the same, that JAMPR significantly outperforms GORT on TW1 and TW3. This is because JAMPR is balancing the cost regarding distance as well as the total duration and incurred waiting time of the tours, while GORT generally rather finds solutions with a smaller total distance but much higher total duration and waiting times. Finally to provide further comparison we here also include a random baseline (\textit{random (1000)} ) which just samples 1000 sequentially created random solutions for each instance and selects the best one.

\begin{table}[h!]
	\caption{Comparison of results for CVRP-TW variants. Best results for each problem are \textbf{bold}. Results of GORT* which are better than the results of the original model are \underline{underlined}.}
	\label{tab:detailed_results}
	\centering
	%\footnotesize
	\small
	\begin{tabular}{llrrrrrrrr}
		\toprule
		& \textbf{Model} & \multicolumn{4}{c}{\textbf{N=20}} & \multicolumn{4}{c}{\textbf{N=50}} \\
		& & \texttt{cost } & \texttt{k } & \texttt{dist } & \texttt{t\textsubscript{inf} } &
			\texttt{cost } & \texttt{k } & \texttt{dist } & \texttt{t\textsubscript{inf} } \\ 
		\hline
		\hline
		\multirow{7}{*}{\rotatebox[origin=c]{90}{\parbox[c]{1cm}{\centering \textbf{TW1}}}} % hard
		& GORT - \textit{AU} 				& 2577.08  & 4.18 & 643.77  & 0.22s & 4344.88  & 6.30 &1110.68 & 1.76s   \\
		& GORT - \textit{GLS} 				& 2522.85  & 4.11 & 617.77  & 8.00s & 4213.23  & 6.15 &1090.36 & 8.06s   \\
		& GORT* - \textit{AU}				& 2873.56  & 5.34 & 909.09  & - & \underline{4025.13}  & 6.74 & 1520.56  & -    \\
		& GORT* - \textit{GLS}				& \underline{2478.70}  & 4.77 & 809.75  & - & \underline{3675.83}  & 6.22 & 1433.05  & -    \\
		& \textit{random (1000)}			& 3036.39 & 5.68 & 1200.37 & - & 7297.53 & 11.80 & 2914.86 & - \\
		& AM\textsuperscript{+TW} (greedy) 	& 3766.88  & 5.29 & 1264.40  & 0.05s & 7189.43  & 9.42 	&2882.53 & 0.12s   \\
		& AM\textsuperscript{+TW} (sampl.) 	& 3041.24  & 5.74 & 1202.64  & 0.12s & 7327.09  & 11.92 &2921.39 & 0.38s   \\
		& AM\textsuperscript{+TW} ($t_{10240}$) & 2750.06  & 5.27 & 1163.58  & 0.95s & 6878.84  & 11.28 &2865.30 & 3.08s   \\
		& JAMPR (greedy) 	 				& 1862.40  & 2.25 & 966.74  & 0.10s & 3055.94  & 5.42 &1733.24 & 0.24s   	\\
		& JAMPR (sampl.) 					& \textbf{1716.60}  & 2.29 & 965.42  & 0.86s & \textbf{2691.55}  & 4.03 &1811.06 & 3.07s   \\
		\hline
		\multirow{7}{*}{\rotatebox[origin=c]{90}{\parbox[c]{1cm}{\centering \textbf{TW2}}}} % wait
		& GORT - \textit{AU} 				& 635.06  & 4.23 & 635.06  & 0.37s & 1123.82  & 6.72 &1123.82 	& 3.81s   \\
		& GORT - \textit{GLS} 				& \textbf{619.57}  & 4.14 & 619.21  & 8.30s & \textit{\textbf{1119.07}}  & 6.67 &1118.01 & 8.09s   \\
		& GORT* - \textit{AU}				& 855.66  & 4.94 & 855.65  & - & 1498.25  & 6.35 & 1498.25  & -    \\
		& GORT* - \textit{GLS}				& 809.01  & 4.76 & 809.01  & - & 1462.33  & 6.27 & 1462.32  & -    \\
		& \textit{random (1000)}			& 1646.83 & 6.13 & 1202.66 & - & 8368.49 & 8.65 & 2897.99 & - \\
		& AM\textsuperscript{+TW} (greedy)	& 7615.69  & 2.00 & 1094.39  & 0.05s & 40245.40  & 2.00 	&2687.95 & 0.12s   \\
		& AM\textsuperscript{+TW} (sampl.) 	& 1572.31  & 6.56 & 1221.09  & 0.11s & 7712.35  & 9.34 	&2953.65 & 0.34s   \\
		& AM\textsuperscript{+TW} ($t_{10240}$) & 1387.30  & 6.46 & 1170.49  & 0.90s & 6730.55  & 9.99 &2954.76& 2.70s    \\
		& JAMPR (greedy) 	 				& 674.72  & 4.32 & 626.80 & 0.11s  & 1273.20  & 6.02 &1126.74 & 0.25s   	\\
		& JAMPR (sampl.) 					& \textit{\textbf{620.68}}  & 4.19 & 602.33  & 0.92s & \textbf{1116.76}  & 5.64 &1076.79 & 2.32s   	\\
		\hline
		\multirow{7}{*}{\rotatebox[origin=c]{90}{\parbox[c]{1cm}{\centering \textbf{TW3}}}} % soft
		& GORT - \textit{AU} 				& 1317.81  & 4.07 	& 637.15  & 1.07s & 2707.72  & 6.12 &1121.57 & 20.63s   	\\
		& GORT - \textit{GLS} 				& 1312.71  & 4.11 	& 625.80  & 8.00s & 2753.66  & 6.18 &1192.61 & 8.02s   	\\
		& GORT* - \textit{AU}				& 1428.53  & 4.74 & 849.69  & - & \underline{2641.83}  & 6.18 & 1480.75  & -    \\
		& GORT* - \textit{GLS}				& 1388.88  & 4.74 & 810.66  & - & \underline{2671.93}  & 6.28 & 1494.14  & -    \\
		& \textit{random (1000)}			& 1409.35 & 3.34 & 953.48 & - & 4407.58 & 6.92 & 2692.78 & - \\
		& AM\textsuperscript{+TW} (greedy) 	& 3101.21  & 2.00 & 1094.39 & 0.05s  & 26467.34  & 2.00 &2687.95& 0.12s    	\\
		& AM\textsuperscript{+TW} (sampl.) 	& 1412.16  & 3.42 & 951.92  & 0.12s & 4161.24  & 7.15 &2674.03 & 0.34s   	\\
		& AM\textsuperscript{+TW} ($t_{10240}$) & 1318.56  & 3.23 & 899.83  & 0.88s & 3941.47  & 6.77  &2575.28 & 2.67s    \\
		& JAMPR (greedy) 	 				& 1002.81  & 1.00 & 733.01  & 0.10s & 3158.26  & 2.01 &1347.72& 0.23s     \\
		& JAMPR (sampl.) 					& \textbf{844.35}  & 1.39 & 660.48  & 0.78s & \textbf{1947.65}  & 2.29 &1358.29 & 2.13s   	\\
		\bottomrule
	\end{tabular}
	
\end{table}

\newpage
\section{Solomon CVRP-TW benchmark}\label{sec:benchmark}
In order to compare the performance of our models on some more established work, we evaluate GORT, JAMPR and AM\textsuperscript{+TW} on two of the available location variants of the Solomon benchmark data set \citep{solomon1987algorithms} which include random (R2) and partially random and clustered (RC2) instances of size 50 and 100 with \textit{hard} TW. 

In the detailed comparison for R201 and RC201 (table \ref{tab:benchmark_detailed}) the N=50 instance includes the depot and the first 50 nodes of the standard N=100 instance. For the aggregated results (table \ref{tab:benchmark_aggregated}) we split the N=100 instances into two instances with 50 nodes, each including the depot and the first and the second half of the nodes respectively.

For the learned models a notable distinction has to be made for the benchmark set for $N=100$. Here we use models which were trained only on the described random instances of size 50 (similar to R201, see section \ref{sec:benchmark_data}). For problem size 100 we evaluate the models which were trained on $N=50$ to investigate the generalization capability similar to the approach in \citep{vinyals_pointer_2015}. Furthermore we test our models on partially clustered data (RC2) which was no explicit part of the training distribution. Additionally the other R2 and RC2 instances vary also in the width of the TW and the number of customers which are constrained by TW (cp. section \ref{sec:benchmark_data}) compared to the training instances.

Again we also include the adapted GORT baseline (explained in section \ref{sec:detailed_exp}) with a waiting time constraint of $\hat{\delta}^k_{a_i}$ equal to 10min (\textit{GORT*}).
Furthermore we give the optimal results (\textit{optimal*}) reported on the benchmark website\footnote{\href{http://web.cba.neu.edu/~msolomon/problems.htm}{http://web.cba.neu.edu/\textasciitilde msolomon/problems.htm}} and the best known heuristic solution\footnote{\href{https://www.sintef.no/projectweb/top/vrptw/solomon-benchmark/100-customers/}{https://www.sintef.no/projectweb/top/vrptw/solomon-benchmark/100-customers/}} (\textit{BKH}). For the aggregated instances we also include the results of \citet{silva2019reinforcement} (\textit{ALS-Q}). Finally we run a random baseline (\textit{random (1000)} ) which just samples 1000 sequentially created random solutions for the instance and selects the best one. 

The results are shown in more detail for the instances R201 and RC201 in table \ref{tab:benchmark_detailed}. Aggregated results for all R2 and RC2 instances respectively are shown in table \ref{tab:benchmark_aggregated}. In general we see that JAMPR is able to sufficiently generalize to instances which are twice the size of the training data set. This can be seen in comparison to the GORT baseline which is run independently on the larger instances. A similar generalization capability is apparent for the results on the partially clustered data (RC2) and different settings for the TW. 

The comparison to other related work is more difficult, since the respective models are normally trained and evaluated on a different objective. Especially the operations research literature focuses on minimizing the number of vehicles as main and the total distance as secondary objective:
\begin{equation}
\operatorname{cost} = \alpha K + \sum^K_{k=1} \left(\sum_{(i,j) \in s_k \uplus\ 0} c_{ij} \right)
\end{equation}
where $\alpha$ is an arbitrary large non-negative penalty factor for the number of vehicles $K$ and $c_{ij}$ is the distance betwenn locations $i$ and $j$.

In contrast we took a more holistic perspective also including waiting time (as well as \textit{soft} TW) into the objective. In this regard we can conclude, that JAMPR finds a good trade-off between transit cost (total distance), waiting times and number of vehicles while taking considerably less inference time than standard heuristic approaches. 
In comparison the sampling behavior of AM\textsuperscript{+TW} is close to random, which means that the model is not able to extract significant features and learn important relations. Greedy inference is better, but still significantly worse than GORT or JAMPR.
Further analysis is provided in section \ref{sec:example_plots}.

\begin{table}[h!t]
	\caption{Detailed results for the R201 and RC201 instance of the Solomon benchmark.}
	\label{tab:benchmark_detailed}
	\centering
	\small
	\begin{tabular}{llrrrrrrrr}
		\toprule
		& \textbf{Model} & \multicolumn{4}{c}{\textbf{N=50}} & \multicolumn{4}{c}{\textbf{N=100}} \\
		& & \texttt{cost } & \texttt{k } & \texttt{dist } & \texttt{t\textsubscript{inf} } 
			& \texttt{cost } & \texttt{k } & \texttt{dist } & \texttt{t\textsubscript{inf} } \\ 
		\midrule
		\multirow{12}{*}{\rotatebox[origin=c]{90}{\parbox[c]{1cm}{\centering \textbf{R201}}}}
		& \textit{optimal*} 	&	-	&	\textit{6.00}	& \textit{791.90}	&	- &	- &	\textit{8.00} &	\textit{1143.20}	&	- \\		
		& \textit{BKH} 			&	-	&	-	&	-	&	-	 &	-	&	\textit{4.00}	&	\textit{1252.37}	&	- \\
		& \textit{random (1000)}		&	\textit{4060.06}	&	\textit{10.00}	&	\textit{2096.71}	&	-	 &	\textit{7953.15}	&	\textit{16.00}	&	\textit{3423.72}	&	- \\
		
		\cmidrule(l{5pt}llr{5pt}){2-10}
		& GORT - \textit{AU} 				&	3807.16	&	7.00	&	822.99	&	0.78	s &	5570.30	&	9.00	&	1235.92	&	3.29s \\		
		& GORT - \textit{GLS} 				&	3696.14	&	7.00	&	804.77	&	8.00	s &	5432.42	&	9.00	&	1223.02	&	8.01s \\
		& GORT* - \textit{AU}				&	1722.53	&	5.00	&	969.36	&	1.12	s &	2785.88	&	7.00	&	1503.86	&	3.80s \\
		& GORT* - \textit{GLS}				&	\textbf{1722.53}	&	5.00	&	969.36	&	8.00	s &	3288.16	&	8.00	&	1455.68	&	8.00s \\
		& AM\textsuperscript{+TW} (greedy) 	&	4468.44	&	8.00	&	1781.85	&	0.21	s &	6604.67	&	12.00	&	3121.32	&	0.31s \\		
		& AM\textsuperscript{+TW} (sampl.) 	&	4694.72	&	11.00	&	1877.49	&	0.49	s &	8924.56	&	20.00	&	3600.92	&	1.20s \\		
		& AM\textsuperscript{+TW} ($t_{10240}$) &	4461.07	&	11.00	&	2050.11	&	3.10	s &	8497.29	&	18.00	&	3327.01	&	8.63s \\	
		& JAMPR (greedy) 	 				&	2279.10	&	4.00	&	1591.65	&	0.34	s &	3436.51	&	6.00	&	2396.45	&	0.59s \\		
		& JAMPR (sampl.) 					&	1791.92	&	\textbf{3.00}	&	1289.35	&	3.08	s &	\textbf{2682.39}	&	\textbf{5.00}	&	2230.59	&	8.98s \\		
		\hline
		\hline
		\multirow{12}{*}{\rotatebox[origin=c]{90}{\parbox[c]{1cm}{\centering \textbf{RC201}}}}
		& \textit{optimal*} 				&	-	&	\textit{5.00}	&	\textit{684.80}	&	-	 &	-	&	\textit{9.00}	&	\textit{1261.80}	&	- \\
		& \textit{BKH}						&	-	&	-	&	-	&	-	 &	-	&	\textit{4.00}	&	\textit{1406.94}	&	- \\
		& \textit{random (1000)}		    &	\textit{5012.32}	&	\textit{13.00}	&	\textit{2579.74}	&	-	 &	\textit{9417.33}	&	\textit{23.00}	&	\textit{4562.85}	&	- \\
		        
		\cmidrule(l{5pt}llr{5pt}){2-10}
		& GORT - \textit{AU} 				&	3249.92	&	6.00	&	688.34	&	0.71	s &	5673.33	&	9.00	&	1443.32	&	3.85s \\
		& GORT - \textit{GLS} 				&	3249.92	&	6.00	&	688.34	&	8.00	s &	6046.57	&	10.00	&	1426.65	&	8.00s \\
		& GORT* - \textit{AU}				&	2968.59	&	8.00	&	1211.88	&	0.51	s &	3188.90	&	7.00	&	1594.82	&	4.67s \\
		& GORT* - \textit{GLS}				&	2649.55	&	7.00	&	1142.04	&	8.00	s &	3188.90	&	7.00	&	1594.82	&	8.00s \\
		& AM\textsuperscript{+TW} (greedy) 	&	4544.19	&	9.00	&	1947.72	&	0.22	s &	7562.65	&	15.00	&	3999.01	&	0.32s \\		
		& AM\textsuperscript{+TW} (sampl.) 	&	5039.80	&	12.00	&	2900.55	&	0.50	s &	9150.92	&	20.00	&	4765.43	&	1.21s \\		
		& AM\textsuperscript{+TW} ($t_{10240}$) &	5000.00	&	13.00	&	2655.76	&	3.13	s &	8633.75	&	18.00	&	4647.16	&	8.72s \\	
		& JAMPR (greedy) 	 				&	1871.79	&	\textbf{3.00}	&	1501.17	&	0.34	s &	3711.35	&	6.00	&	3053.83	&	0.58s \\		
		& JAMPR (sampl.) 					&	\textbf{1819.27}	&	4.00	&	1250.00	&	3.15	s &	\textbf{2904.33}	&	\textbf{6.00}	&	2523.61	&	8.74s \\	
		\bottomrule
	\end{tabular}
	
\end{table}

\begin{table}[h!t]
	\caption{Aggregated average results over all R2 and RC2 instances of the Solomon benchmark}
	\label{tab:benchmark_aggregated}
	\centering
	\small
	\begin{tabular}{llrrrrrrrr}
		\toprule
		& \textbf{Model} & \multicolumn{4}{c}{\textbf{N=50}} & \multicolumn{4}{c}{\textbf{N=100}} \\
		& & \texttt{cost } & \texttt{k } & \texttt{dist } & \texttt{t\textsubscript{inf} } 
		& \texttt{cost } & \texttt{k } & \texttt{dist } & \texttt{t\textsubscript{inf} } \\ 
		\midrule
		\multirow{12}{*}{\rotatebox[origin=c]{90}{\parbox[c]{1cm}{\centering \textbf{R2}}}}
		& \textit{BKH} 						&	-	&	-	&	-	&	-	 &	-	&	\textit{2.73}	&	\textit{951.03}	&	- \\
		& \textit{ALS-Q} 					&	-	&	-	&	-	&	-	 &	-	&	\textit{3.00}	&	\textit{956.61}	&	- \\
		%& \textit{PHGA} 					&	-	&	-	&	-	&	-	 &	-	&	\textit{2.73}	&	\textit{1056.59}	&	\textit{1800.00s} \\
		& \textit{random (1000)}			&	\textit{3235.21}	&	\textit{7.00}	&	\textit{1917.63}	&	-	 &	\textit{5855.62}	&	\textit{14.00}	&	\textit{3468.93}	&	- \\
				
		\cmidrule(l{5pt}llr{5pt}){2-10}
		& GORT - \textit{AU} 				&	2920.98	&	5.82	&	633.09	&	0.95	s &	4519.85	&	8.27	&	987.36	&	5.41s \\
		& GORT - \textit{GLS} 				&	2763.58	&	5.77	&	610.45	&	8.00	s &	4468.01	&	8.18	&	977.93	&	8.00s \\
		& GORT* - \textit{AU}				&	2284.14	&	5.55	&	731.05	&	0.99	s &	2428.52	&	6.09	&	1091.08	&	5.78s \\
		& GORT* - \textit{GLS}				&	2231.35	&	5.50	&	704.01	&	8.00	s &	2476.82	&	6.18	&	1082.11	&	8.00s \\
		& AM\textsuperscript{+TW} (greedy) 	&	3451.73	&	5.64	&	1556.63	&	0.11	s &	6149.99	&	9.36	&	3312.46	&	0.22s \\
		& AM\textsuperscript{+TW} (sampl.) 	&	3287.95	&	7.82	&	1737.51	&	0.35	s &	6407.59	&	12.73	&	3448.46	&	1.00s \\
		& AM\textsuperscript{+TW} ($t_{10240}$) &	2997.14	&	7.59	&	1700.02	&	2.70	s &	6009.37	&	12.27	&	3377.40	&	7.75s \\
		& JAMPR (greedy) 	 				&	1694.49	&	\textbf{2.95}	&	1106.25	&	0.24	s &	2536.46	&	\textbf{4.27}	&	1959.64	&	0.49s \\
		& JAMPR (sampl.) 					&	\textbf{1566.42}	&	3.59	&	1179.40	&	3.04	s &	\textbf{2401.86}	&	5.00	&	2068.57	&	8.85s \\
		\hline
		\hline
		\multirow{12}{*}{\rotatebox[origin=c]{90}{\parbox[c]{1cm}{\centering \textbf{RC2}}}}
		& \textit{BKH} 						&	-	&	-	&	-	&	-	 &	-	&	\textit{3.25}	&	\textit{1119.24}	&	- \\
		& \textit{ALS-Q} 					&	-	&	-	&	-	&	-	 &	-	&	\textit{3.38}	&	\textit{1164.61}	&	- \\
		%& \textit{PHGA} 					&	-	&	-	&	-	&	-	 &	-	&	\textit{3.25}	&	\textit{1258.15}	&	\textit{1800.00s} \\
		& \textit{random (1000)}			&	\textit{4833.41}	&	\textit{12.00}	&	\textit{2691.28}	&	-	 &	\textit{8696.28}	&	\textit{19.00}	&	\textit{4800.89}	&	- \\
				
		\cmidrule(l{5pt}llr{5pt}){2-10}
		& GORT - \textit{AU} 				&	2794.08	&	5.56	&	618.30	&	0.89	s &	4477.10	&	8.38	&	1127.18	&	4.83s \\
		& GORT - \textit{GLS} 				&	2756.02	&	5.56	&	602.51	&	8.00	s &	4535.50	&	8.50	&	1123.41	&	8.00s \\
		& GORT* - \textit{AU}				&	2244.97	&	5.63	&	837.89	&	0.87	s &	2981.47	&	6.88	&	1278.96	&	5.35s \\
		& GORT* - \textit{GLS}				&	2240.78	&	5.75	&	791.05	&	8.00	s &	2971.08	&	6.88	&	1274.59	&	8.00s \\
		& AM\textsuperscript{+TW} (greedy) 	&	3865.95	&	6.38	&	1754.63	&	0.11	s &	7015.59	&	11.38	&	3877.71	&	0.23s \\
		& AM\textsuperscript{+TW} (sampl.) 	&	3914.12	&	8.94	&	2175.67	&	0.36	s &	7632.03	&	15.25	&	4372.87	&	1.03s \\
		& AM\textsuperscript{+TW} ($t_{10240}$) &	3601.40	&	8.69	&	2102.26	&	2.75	s &	7363.07	&	15.00	&	4356.63	&	8.03s \\
		& JAMPR (greedy) 	 				&	1766.32	&	\textbf{3.06}	&	1174.98	&	0.25	s &	3101.57	&	5.00	&	2484.69	&	0.50s \\
		& JAMPR (sampl.) 					&	\textbf{1645.00}	&	3.25	&	1207.42	&	3.05	s &	\textbf{2774.64}	&	\textbf{5.00}	&	2401.82	&	9.01s \\
		\bottomrule
	\end{tabular}
	
\end{table}

\clearpage
\section{Example solution plots}\label{sec:example_plots}
We plot the solutions to the R201 and RC201 instances for both problem sizes ($N=50$ and $N=100$) always for the best variant of each of the three compared methods GORT, AM\textsuperscript{+TW} and JAMPR. The letters in the legend describe for each tour the number of customer (n), the length (l) i.e.\ total distance, the used capacity (q) and the total duration (t). Plots are best viewed in color. 
We want to remind the reader that these instances are similar to the TW1 variant with \textit{hard} time windows and therefore tours are not only dependent on the geographical proximity, leading to tours which are visually less clearly separated.

\begin{figure}[h!]
	\centering
	\includegraphics[width=0.49\textwidth,valign=t]{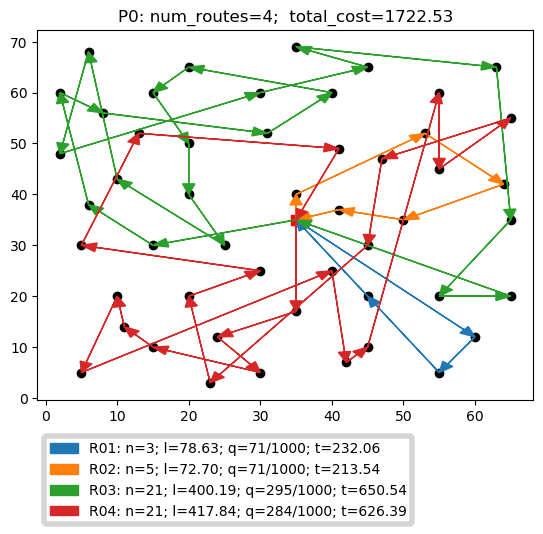}
	\includegraphics[width=0.49\textwidth,valign=t]{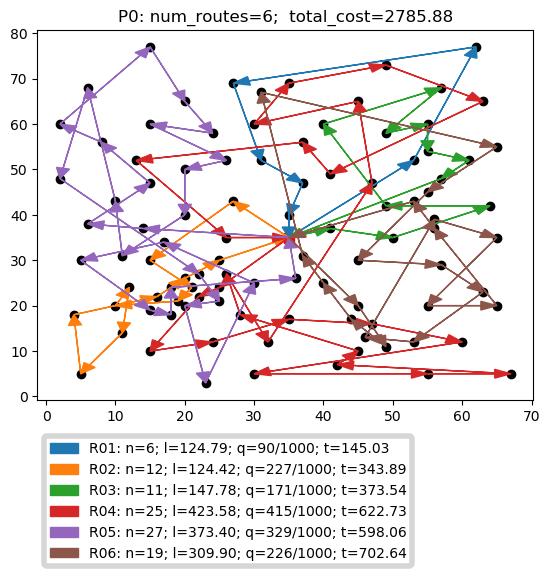}
	\caption{GORT solution for R201-50 (left) and R201-100 (right).}
\end{figure}
\begin{figure}[h!]
	\centering
	\includegraphics[width=0.49\textwidth,valign=t]{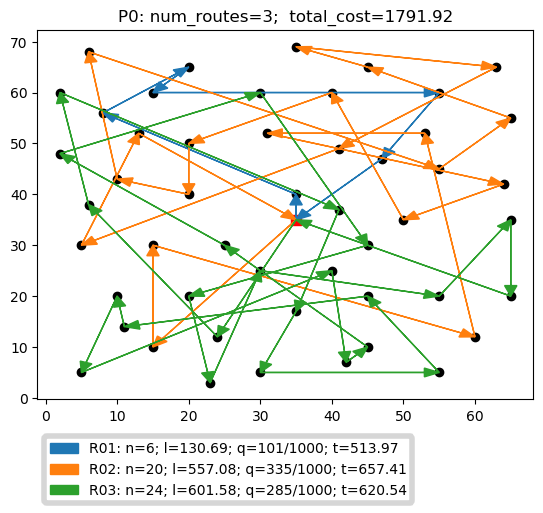}
	\includegraphics[width=0.49\textwidth,valign=t]{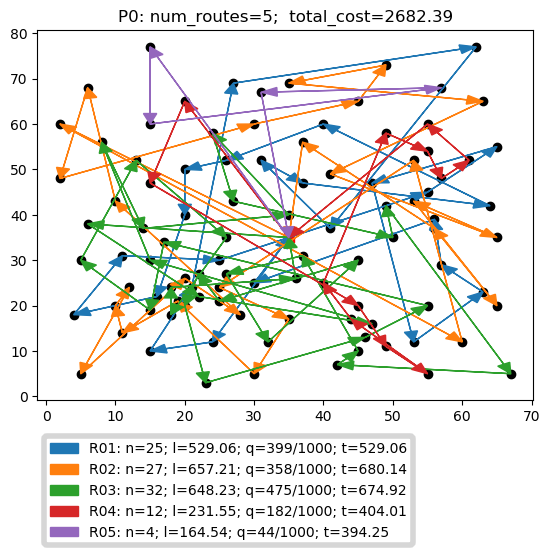}
	\caption{JAMPR solution for R201-50 (left) and R201-100 (right).}
\end{figure}
\begin{figure}[h!]
	\centering
	\includegraphics[width=0.49\textwidth,valign=t]{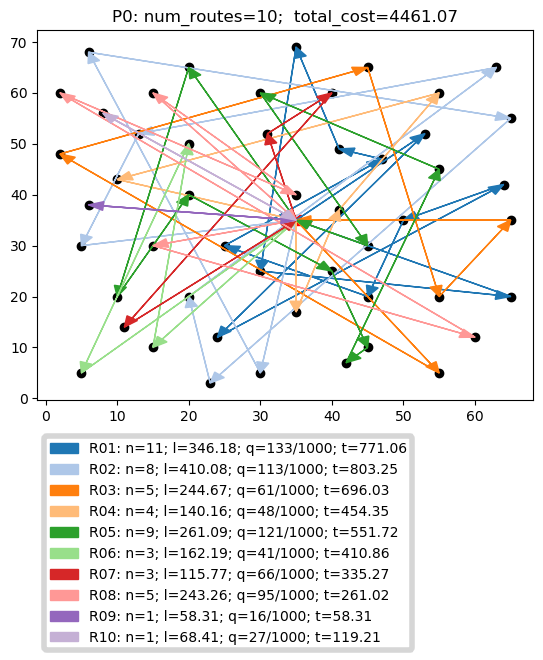}
	\includegraphics[width=0.49\textwidth,valign=t]{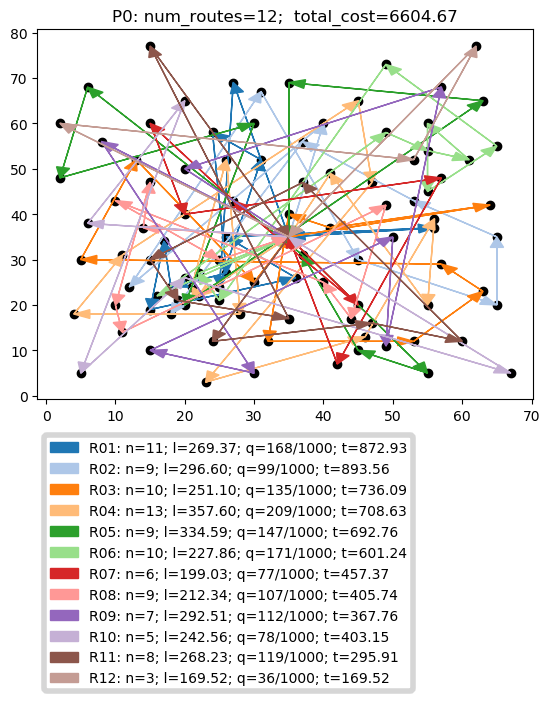}
	\caption{AM\textsuperscript{+TW} solution for R201-50 (left) and R201-100 (right).}
\end{figure}

\begin{figure}[h!]
	\centering
	\includegraphics[width=0.49\textwidth,valign=t]{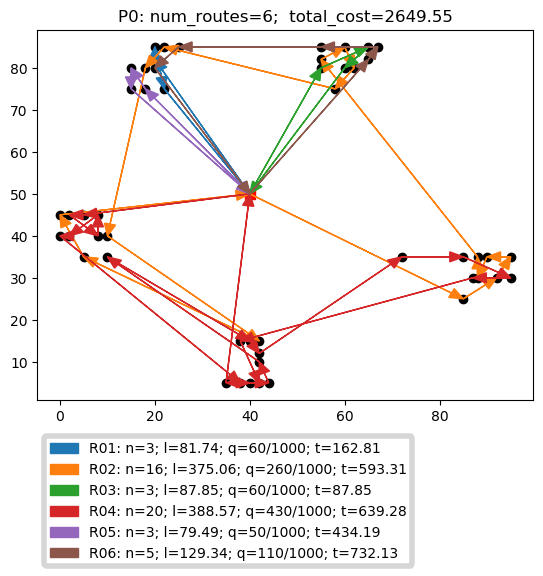}
	\includegraphics[width=0.49\textwidth,valign=t]{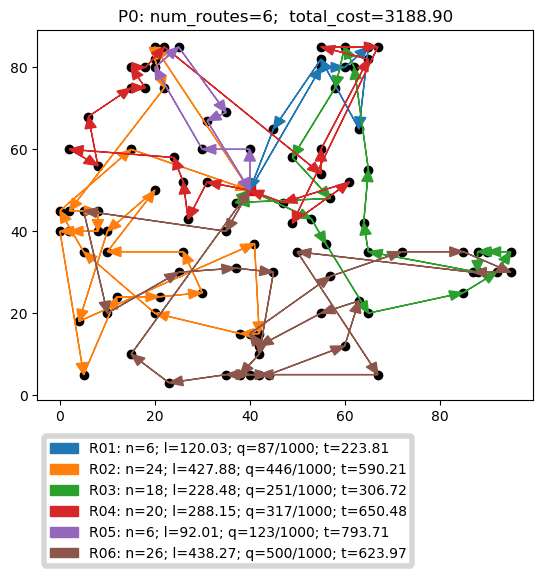}
	\caption{GORT solution for RC201-50 (left) and RC201-100 (right).}
\end{figure}
\begin{figure}[h!]
	\centering
	\includegraphics[width=0.49\textwidth,valign=t]{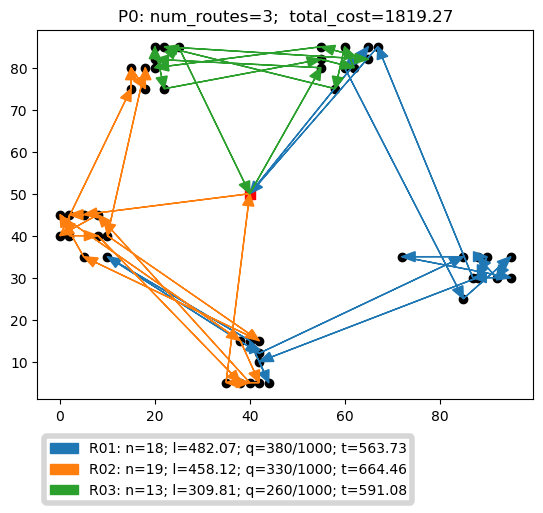}
	\includegraphics[width=0.49\textwidth,valign=t]{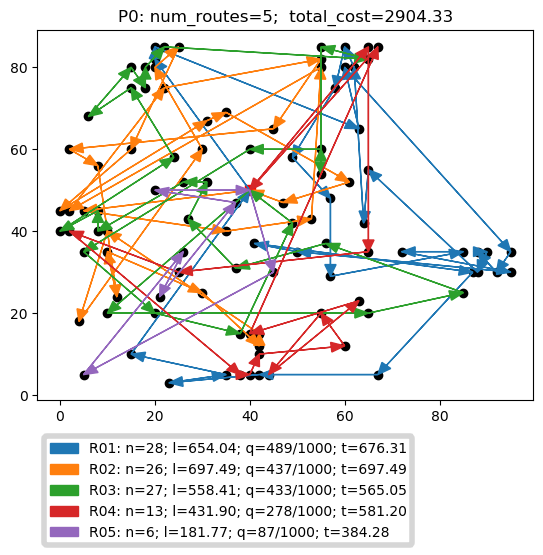}
	\caption{JAMPR solution for RC201-50 (left) and RC201-100 (right).}
\end{figure}
\begin{figure}[h!]
	\centering
	\includegraphics[width=0.49\textwidth,valign=t]{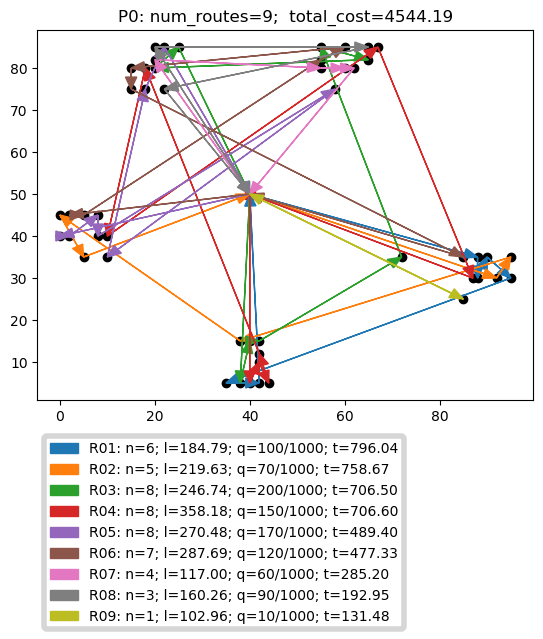}
	\includegraphics[width=0.49\textwidth,valign=t]{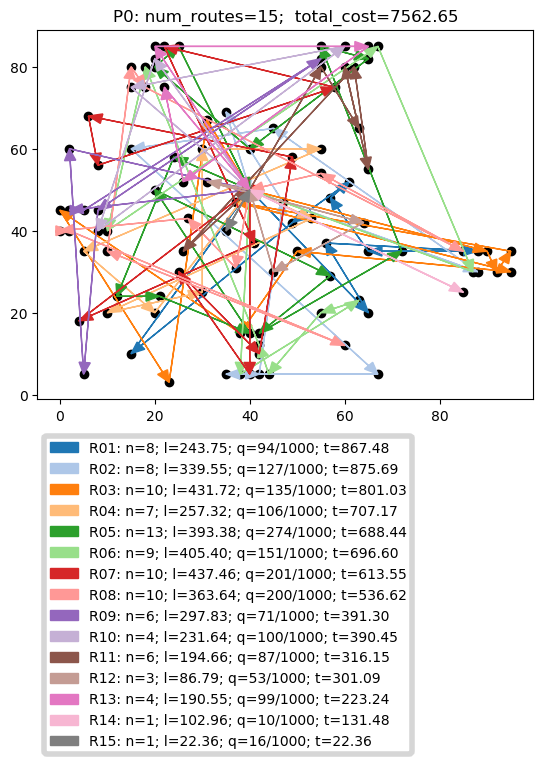}
	\caption{AM\textsuperscript{+TW} solution for RC201-50 (left) and RC201-100 (right).}
	\label{fig:rc201_am}
\end{figure}

As we can see from the solution plots, the tours of GORT are based more on geographical proximity, leading to tours with smaller total distance but a higher number of used vehicles with less used capacity and much more waiting times leading to longer tour duration. In contrast the tours created by JAMPR focus less on close neighbors and more on most efficiently using the available time and capacity, leading to tours with larger total distance but a smaller number of used vehicles with a tighter schedule. This generally leads to a shorter total duration. 
Finally AM\textsuperscript{+TW} does not seem able to focus on any part of the objective and constraint space with close to random solutions. This can be seen especially in figure \ref{fig:rc201_am} for the partially clustered locations, where on the $N=50$ instance e.g.\ tour R04 erratically visits customers in all clusters.

\clearpage
\section{Effect of max concurrency}

\begin{figure}[h!]
	\centering
	\includegraphics[width=1.0\textwidth]{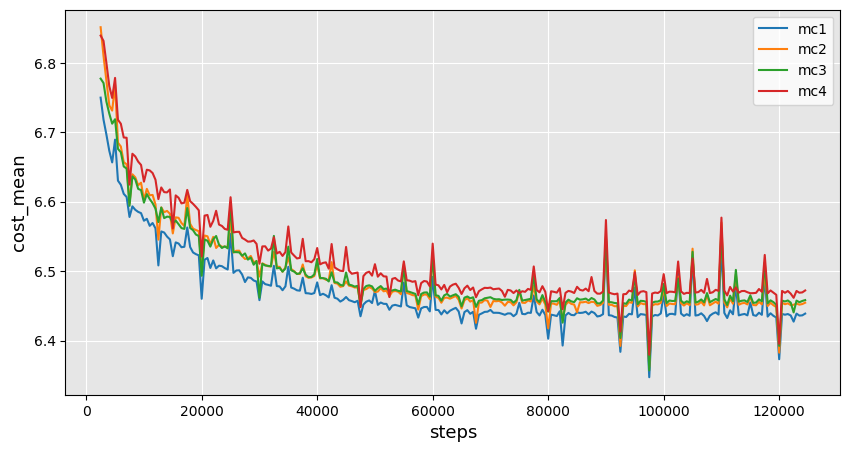}
	\caption{
		Learning curves on random sampled training data for the CVRP of size $N=20$.
	}
	\label{fig:lc-CVRP-20}
\end{figure}

\begin{figure}[h!]
	\centering
	\includegraphics[width=1.0\textwidth]{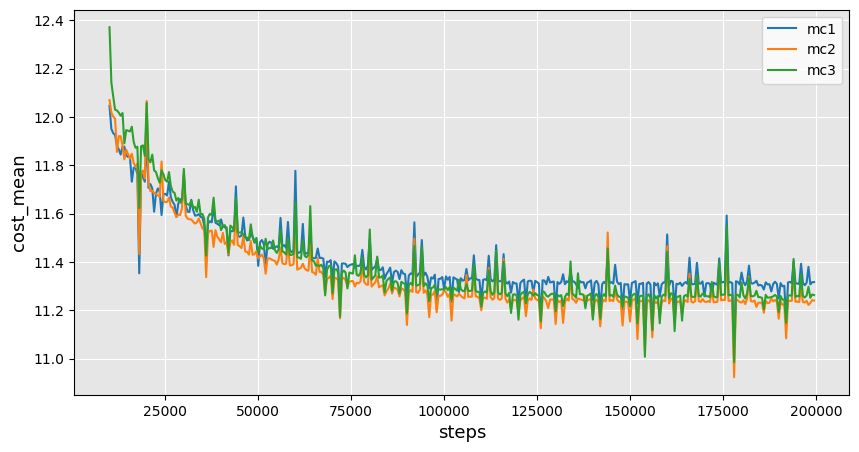}
	\caption{
		Learning curves on random sampled training data for the CVRP of size $N=50$.
	}
	\label{fig:lc-CVRP-50}
\end{figure}

Finally we show some learning curves for the CVRP-TW variant TW1 with hard time windows and the standard CVRP for different values of maximum concurrency controlled by $m_{\text{con}}$ (mc). For the less constrained CVRP a smaller number of $m_{\text{con}}=1$ (N=20, figure \ref{fig:lc-CVRP-20}) or $m_{\text{con}}=2$ 
(N=50, figure \ref{fig:lc-CVRP-50}) works best. In contrast the much more constrained CVRP-TW1 benefits from a higher $m_{\text{con}}$. However, since most of the problem instances are solved by JAMPR with two to four vehicles on average, a larger $m_{\text{con}}$ does not have any additional advantage, but rather hurts performance by unnecessarily increasing learning complexity.

\begin{figure}[h!]
	\centering
	\includegraphics[width=1.0\textwidth]{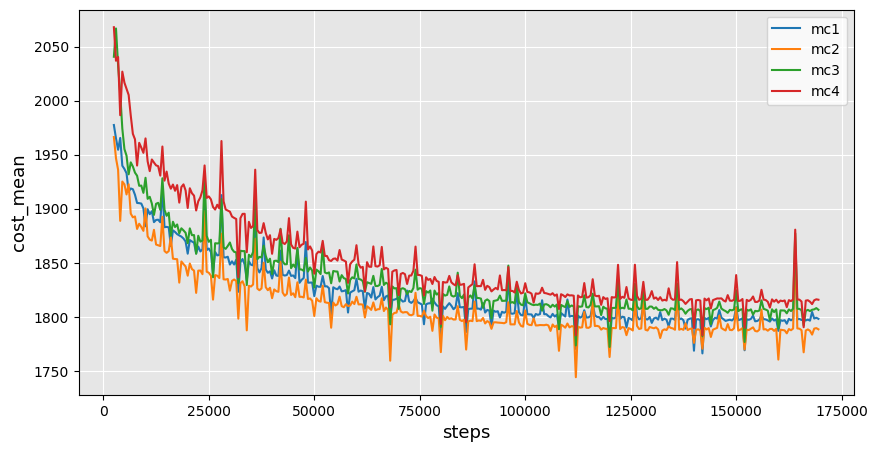}
	\caption{
		Learning curves on random sampled training data for the CVRP-TW1 of size $N=20$.
	}
	\label{fig:lc-CVRP-TW1-20}
\end{figure}

\begin{figure}[h!]
	\centering
	\includegraphics[width=1.0\textwidth]{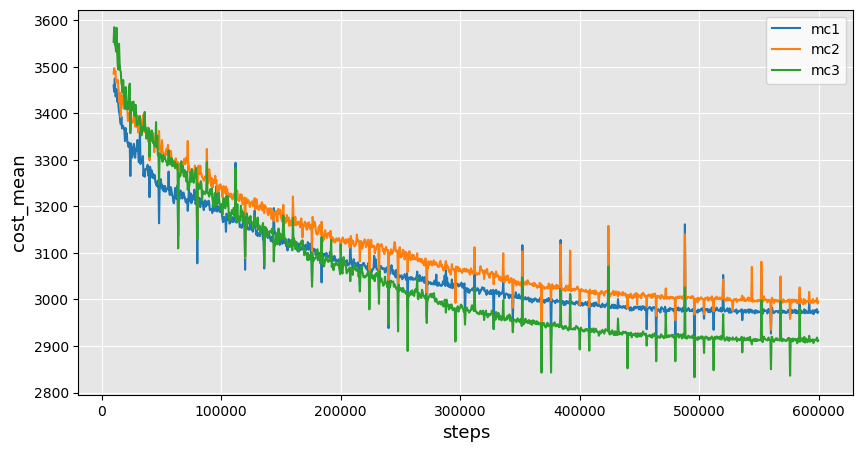}
	\caption{
		Learning curves on random sampled training data for the CVRP-TW1 of size $N=50$.
	}
	\label{fig:lc-CVRP-TW1-50}
\end{figure}

\clearpage

\end{document}